\pdfoutput=1

\documentclass[11pt]{article}

\usepackage[final]{acl}

\usepackage{times}
\usepackage{latexsym}

\usepackage[T1]{fontenc}

\usepackage[utf8]{inputenc}

\usepackage{microtype}

\usepackage{inconsolata}

\usepackage{amsmath}
\usepackage{graphicx}
\usepackage{array}     
\usepackage{booktabs}  
\usepackage{amssymb}   
\usepackage{wasysym}   

\usepackage[capitalize]{cleveref}
\crefname{section}{Sec.}{Secs.}
\Crefname{section}{Section}{Sections}
\Crefname{table}{Table}{Tables}
\crefname{table}{Tab.}{Tabs.}
\crefname{appendix}{Sec.}{Secs.}
\Crefname{appendix}{Section}{Sections}

\usepackage[most]{tcolorbox} 

\usepackage{multirow}
\usepackage{booktabs}
\usepackage{adjustbox}

\usepackage{xspace}
\usepackage{subcaption}

\newcommand{\model}{\textsc{MEXA}\xspace}

\title{\texorpdfstring{\includegraphics[width=0.04\linewidth]{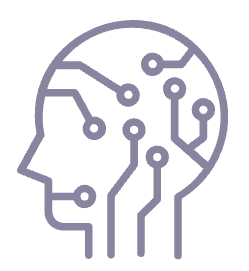}}\textsc{MEXA}: Towards General Multimodal Reasoning with\\Dynamic Multi-Expert Aggregation}

\author{%
Shoubin Yu$^{1,}$\thanks{Equal contribution.} \quad 
Yue Zhang$^{1,*}$ \quad 
Ziyang Wang$^{1}$ \quad \\
\textbf{Jaehong Yoon}$^{1,2}$ \quad  
\textbf{Mohit Bansal}$^{1}$ \\\\
$^{1}$UNC Chapel Hill \quad\quad $^{2}$Nanyang Technological University\\
\\
{{ \tt \normalsize \href{https://github.com/Yui010206/MEXA}{\textcolor{magenta}{https://github.com/Yui010206/MEXA}} }}}

\begin{document}
\maketitle

\begin{abstract}
Combining pre-trained expert models offers substantial potential for scalable multimodal reasoning, but building a unified framework remains challenging due to the increasing diversity of input modalities and task complexity.
For instance, medical diagnosis requires precise reasoning over structured clinical tables, while financial forecasting depends on interpreting plot-based data to make informed predictions.
To tackle this challenge, we introduce \model{}, a training-free framework that performs modality- and task-aware aggregation of multiple expert models to enable effective multimodal reasoning across diverse and distinct domains.
\model{} dynamically selects expert models based on the input modality and the task-specific reasoning demands (i.e., skills). 
Each expert model, specialized in a modality-task pair, generates interpretable textual reasoning outputs. 
\model{} then aggregates and reasons over these outputs using a Large Reasoning Model (LRM) to produce the final answer.
This modular design allows flexible and transparent multimodal reasoning across diverse domains without additional training overhead.
We extensively evaluate our approach on diverse multimodal benchmarks, including Video Reasoning, Audio Reasoning, 3D Understanding, and Medical QA. \model{} consistently delivers performance improvements over strong multimodal baselines, highlighting the effectiveness and broad applicability of our expert-driven selection and aggregation in diverse multimodal reasoning tasks.
\end{abstract} 
\section{Introduction}

The rapid advancement of multimodal learning has significantly improved AI systems' ability to understand, reason, and interact with the real world, benefiting diverse tasks such as visual question answering~\cite{antol2015vqa,hudson2019gqa,marino2019ok,mathew2021docvqa,tanaka2023slidevqa,winata2024worldcuisines, wang2025videotree, zhang2024common, guo2025rethinking}, medical image diagnosis~\cite{lau2018dataset,he2020pathvqa,liu2021slake,azam2022review,cai2023pre,bai2024m3d}, and embodied AI~\cite{brohan2023rt,driess2023palm,liu2024aligning, durante2024agent,majumdar2024openeqa, zhang2024vision}.
As AI systems become more integrated into complex, real-world applications, the ability to flexibly understand and reason upon heterogeneous multimodal inputs is increasingly essential. 
For example, a medical diagnostic system may need to interpret CT scans, extract structured information from clinical notes, and reason about temporal patterns in patient histories, while a financial forecasting system must effectively analyze and interpret plot data to make informed predictions about market trends and economic risks.

However, despite recent advances in multimodal learning, the increasing diversity and complexity of multimodal data pose significant challenges for developing a flexible and unified framework that can reason effectively across modalities and generalize across tasks requiring diverse skills.
Existing multimodal architectures typically require training separate encoders tailored to each modality, along with designing complex cross-modal alignment mechanisms~\cite{tan2019lxmert, li2022blip, li2023blip, liu2023visual, hong20233d, yu2024crema}. 
While effective, these models require fine-tuning for each specific task, leading to substantial training overhead and limiting their adaptability to new modalities or tasks.
Moreover, previous approaches~\cite{jaegle2021perceiver,team2024chameleon} often implicitly fuse multimodal inputs at early stages, restricting transparency and interpretability of the reasoning process. Instead, an ideal multimodal framework should seamlessly process inputs from any modality for any given task, dynamically directing these inputs to the most suitable multimodal expert models. 
This necessitates a modular design where each expert model is associated explicitly with skills corresponding to particular modality-task, thereby ensuring precise skill matching and enhancing interpretability.

In this paper, we propose \textbf{M}ultimodal \textbf{Ex}pert \textbf{A}ggregator~(\model{}), a novel \textbf{training-free multi-expert aggregation framework} that dynamically coordinates a pool of specialized experts. \model{} selectively activates and aggregates reasoning outputs from the most relevant expert models for each input instance, guided by both the input modalities (\textit{e.g.}, image, audio, 3D, medical scan) and the required level of reasoning demands~(i.e., skills) for that instance.
\model{} spans a wide spectrum of tasks, from low-level perceptual reasoning (\textit{e.g.}, object recognition or OCR) to high-level cognitive inference (\textit{e.g.}, temporal event understanding in video or diagnostic interpretation in medical imaging).
Unlike existing methods that rely on monolithic, end-to-end fine-tuned multimodal models or fixed, statically defined expert pipelines, \model{}’s modular expert coordination enables flexible, interpretable, and scalable multimodal reasoning.

Specifically, we first assemble a diverse pool of expert models, each designed to handle distinct aspects of modalities and tasks, to address mainstream multimodal challenges effectively. Each expert specializes in extracting information unique to its respective skill and encoding it into a unified textual representation. 
Then, to effectively coordinate these experts, we design an expert selection module that employs a Multimodal Large Language Model (MLLM) as a versatile router within our framework. 
This router dynamically selects and activates the appropriate experts by analyzing the modality of the input data and identifying the specific skills required based on the input query and task requirements. 
Finally, we introduce an aggregator that reasons over the outputs of the selected experts using a Large Reasoning Model (LRM), which excels at long-context understanding and producing complex long CoT reasoning during LRM inference. Instead of relying on heuristic merging, the aggregator systematically integrates the complementary information from each expert and infers the final answer based on the full context of the target task.

\begin{figure*}[t]
    \centering
    \includegraphics[width=\linewidth]{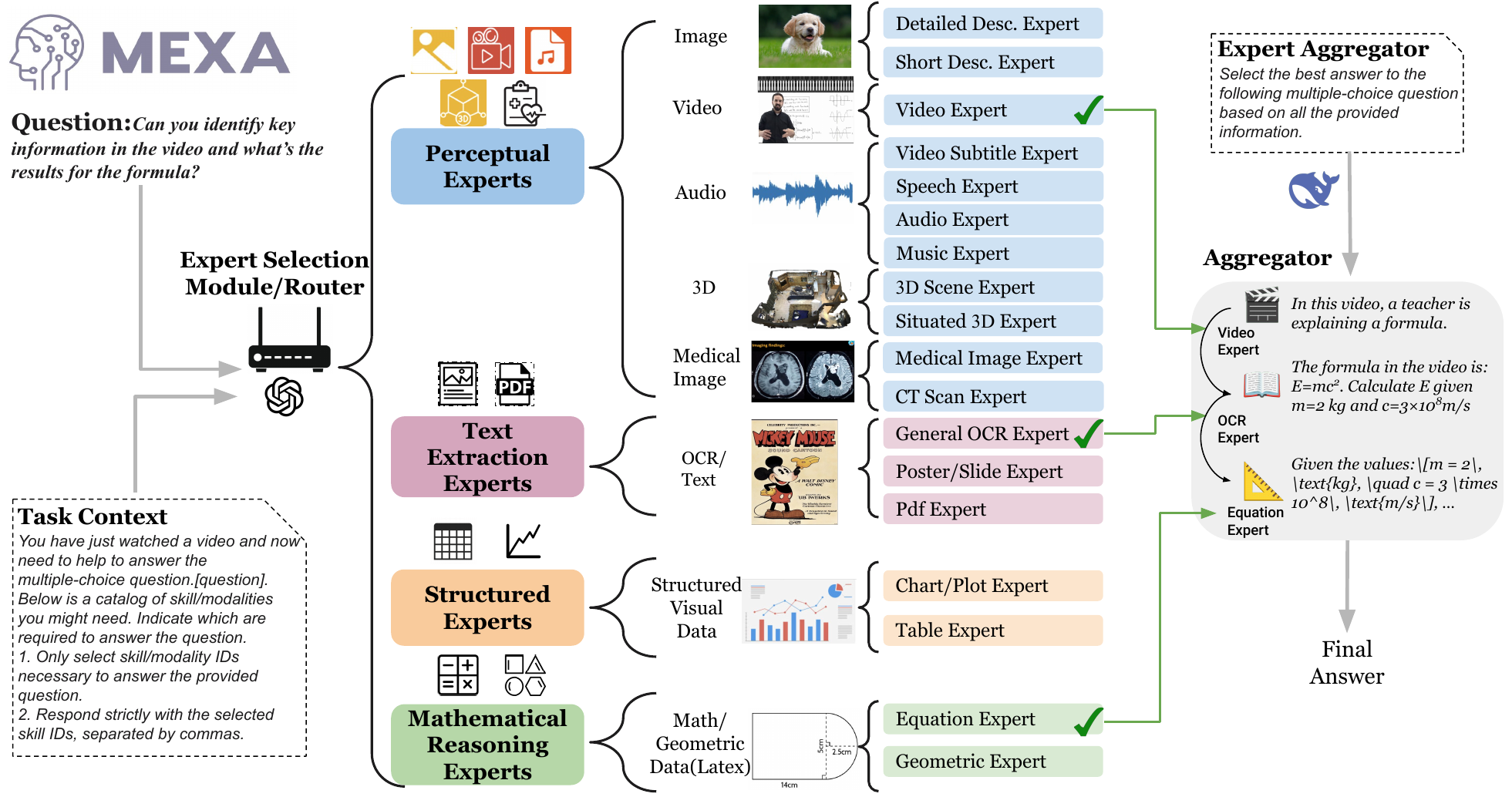}
    \caption{
    Overview of the \model{} Architecture. Given the input task context and question, \model{} first employs an MLLM router~(\cref{MLLM Router}) to select the appropriate experts based on input modality and required reasoning skills. The aggregator~(\cref{aggregator}) then reasons over the outputs from the selected experts to generate the final answer.}
    \label{fig:MEXA architecture}
\end{figure*}


We perform extensive evaluations of our framework on a diverse set of challenging multimodal benchmarks containing various modalities and requiring diverse skills, including expert knowledge heavy video reasoning (Video-MMMU~\cite{hu2025video}), audio QA (MMAU~\cite{sakshi2024mmaumassivemultitaskaudio}), 3D scene understanding (SQA3D~\cite{ma2022sqa3d}), and medical imaging-based QA (M3D~\cite{bai2024m3d}).  
Experimental results show that our proposed method consistently outperforms strong multimodal baseline models across all evaluated benchmarks, with accuracy gains of +5.7\% on Video-MMMU, +12.2\% on MMAU, +1.7\% on SQA3D, and +1.6\% on M3D in the corresponding metrics, demonstrating the effectiveness, flexibility, and robustness of our expert-driven multimodal aggregation framework.
We further conduct ablation studies on expert-selection router design and LLM aggregator
to provide more insights for future work. Our contributions are summarized as follows: 

\begin{itemize}
    \item We present \model{}, a training-free, modality-extensible, and flexible framework that handles general multimodal reasoning via dynamic expert selection and aggregation. 
    \item \model{} achieves strong performance on challenging multimodal benchmarks across diverse modalities, against specialized models as a more general framework. 
    \item We conduct ablation studies and analyses based on our \model{} framework, providing insights into the design choices and effectiveness of each component.  
\end{itemize}


\section{Related Work}

\paragraph{Mixture of Multiple Expert Models.} 
Recently, the Mixture-of-Experts (MoE) paradigm~\cite{shazeer2017outrageously, zoph2022st, chen2023mod, zhou2022mixture, dai2022stablemoe,zhang2023robust,chowdhury2023patch,lee2024becotta}, which integrates multiple specialized parametric \textit{expert modules}, has been widely adopted across a range of machine learning tasks to leverage complementary capabilities. These methods typically employ sparsely-gated mechanisms that activate only a subset of expert modules during each forward pass, improving computational efficiency and scalability.
Building on this motivation, recent advances have explored leveraging a mixture of \textit{expert models} (or \textit{agents})~\citep{li2024smoa, wang2024mixture, chen2025symbolic,li2025rethinking, cao2025multi2} that combine independently pre-trained models across diverse knowledge domains, moving beyond the traditional layer-wise sparse activation used in MoE-based approaches. Rather than operating as interchangeable sub-modules within a single model, such agent-based (or model-based) mixtures selectively aggregate the outputs of independently trained models through dynamic routing mechanisms to solve tasks. This design allows for more flexible and targeted knowledge utilization, promoting improved generalization across tasks and greater extensibility.
However, existing approaches primarily focus on single-modality settings~\cite{chen2025symbolic} or lack support for complex reasoning tasks~\cite{li2024smoa, wang2024mixture, li2025rethinking, cao2025multi2, zhang2025silvr}. In contrast, our method extends this line of work by scaling expert modularity to handle heterogeneous multimodal inputs and reasoning, enabling more versatile and adaptive problem-solving across domains.

\paragraph{Many-Modal Understanding and Reasoning.}
Real-world environments are increasingly dynamic, requiring the need for AI systems to perceive and process a broader range of modalities beyond unimodal learning.
This growing demand has led to advances in models that integrate diverse signals such as images, audio, text, and 3D point clouds to improve learning and generalization. 
Among these, Vision-Language Models (VLMs)~\citep{huang2022large,li2023otter,gong2023multimodalgpt,chen2024vast} support tasks like speech recognition and audio captioning by fusing acoustic and textual information. Similarly, 2D-3D Joint Models~\citep{li2020detailed,hou2021pri3d,hou2023mask3d,lei2024vit} combine spatial and geometric features for enhanced 3D scene understanding. However, these models are typically limited to fixed modality-task combinations and struggle to generalize or scale to novel modality inputs. 
This motivates the development of flexible many-modality systems~\citep{zellers2022merlot,han2023onellm,li2023blip,girdhar2023imagebind,liu2023prismer,yu2024crema} that adapt to heterogeneous inputs and reasoning demands across domains. 
Yet, most existing systems are primarily designed to fuse multiple input modalities and learn perception and reasoning implicitly within a unified architecture. While effective for many tasks, these approaches often struggle with scalability and tend to overlook the explicit use of LLMs' advanced reasoning capabilities.
Our modular framework selects and coordinates expert models through an LLM-driven router and aggregator, directly leveraging reasoning abilities that are insufficiently exploited by unified multimodal architectures.
Another line of work, programming-based methods~\cite{lu2023chameleon,suris2023vipergpt}, relies on pre-defined functions or in-context examples (few-shot prompts) and executes reasoning in a code format. 
In contrast,  our MEXA does not require in-context examples and operates entirely in a training-free, zero-shot setting.  MEXA performs deep reasoning using a Large Reasoning Model rather than code execution, allowing for free-form input rather than rigid programmatic steps. This flexibility makes MEXA more extensible and generalizable to a wider range of tasks and modalities, including audio, 3D, and medical tasks, which are beyond the scope of ViperGPT~\cite{suris2023vipergpt} and Chameleon~\cite{lu2023chameleon}.

\section{\model{}: Dynamic Multi-Modal Expert Aggregation}
We first highlight the challenges faced by general-purpose multimodal reasoning systems in~\cref{general purpose} and introduce our skill-specialized expert aggregation framework as a solution in~\cref{our method}. Within~\cref{our method},  we first describe the organization of a pool of expert modules and our strategy for extracting expertise-specific text representations from the specialized experts(\cref{expert model pool}).
Next, we introduce an expert selection module that dynamically determines which experts to activate based on the input query, target modality, required skill, and reasoning complexity~(\cref{MLLM Router}). Finally, we present our expert aggregation strategy, which integrates the outputs from the selected expert to generate a final answer~(\cref{aggregator}).

\subsection{General-Purpose Multimodal Reasoning}
\label{general purpose}
Existing general-purpose multimodal reasoning systems~\cite{han2023onellm,liu2023prismer,yu2024crema} typically rely on fixed fusion architectures that process raw multimodal inputs within a shared representation space. These systems aim to implicitly learn both multimodal alignment and reasoning capabilities within a single end-to-end model. 
Formally, such a system can be framed as a question-answering model that takes a natural language question \( Q_{\text{ques}} \) and a collection of supported modalities, such as 2D images, 3D point clouds, videos, text, and audio, denoted as \( \mathcal{M}= \{m_1, m_2, \dots, m_n\} \), where $n$ is the number of input modalities as input. The input modalities are first encoded and fused into a joint representation space, which is then processed by a monolithic reasoning model trained to reason over this fused representation. The model produces a contextually grounded answer $A$, defined as:
\begin{equation}
    A = f_{\text{fuse}}(f_\text{enc}(\mathcal{M}), Q_{\text{ques}}),
\label{general equation}
\end{equation}
where\( f_{\text{enc}}(\cdot) \) denotes the encoding and fusion of the multimodal inputs \( \mathcal{M} \), and \( f_{\text{fuse}}(\cdot) \) 
represents the general-purpose reasoning function applied to the fused multimodal representation to generate the answer.
While this approach offers architectural simplicity, it often struggles with scalability, interpretability, and adaptability, especially when dealing with diverse input modalities and varying reasoning demands. This highlights the need for a novel multimodal reasoning framework capable of seamlessly processing inputs across modalities and dynamically activating the appropriate skills for different reasoning tasks.

\subsection{Skill-Specialized Mixture of Expert Models}
\label{our method}
To overcome the limitations of existing general-purpose multimodal models in handling diverse modalities and tasks, we propose \model{}, a flexible and expert-driven multimodal reasoning framework. Instead of processing all modality inputs uniformly through fixed architectures, \model{} dynamically routes queries to a set of specialized expert models, selected based on input modality and task complexity.
Formally, our system takes a question $Q_\text{ques}$ as input and generates an answer $A$ in response. 
To accomplish this, we first employ an expert selection module that serves as a router~(denoted as $\texttt{Router}$), which selects a subset of expert models, denoted as $E_s$. The outputs from the selected experts are then passed to a reasoning module, functioning as an aggregator~(denoted as $\texttt{Aggregator}$), which integrates and reasons over the information extracted by expert models to generate the final answer $A$: 
\begin{equation}
\begin{split}
    A = \texttt{Aggregator}( \{ E_s\mid E_i\in
    \texttt{Router}(
    \\Q_\text{ques}, \cdots) \}).
\end{split}
\end{equation}

In contrast to Equation~\ref{general equation}, which implicitly fuses all modalities within a single architecture, our method enables specialization and modular design. It supports dynamic adaptation by selecting relevant experts based on the task, where each expert plays a well-defined role.

\subsubsection{Design Principles for the Expert Pool} 
\label{expert model pool}
Our expert pool is designed based on the following key principles.

\paragraph{Design Principle 1: Task-Aware and Modality-Sensitive Reasoning. }
To ensure that our expert pool is both modular and broadly applicable, we construct it by analyzing the modalities and skills commonly required across diverse multimodal tasks. While our approach is designed to be task-agnostic and extensible, we use the \textit{Video-MMMU} dataset~\cite{hu2025video} as a representative case study to guide the initial design and evaluation. 
This benchmark includes videos from a wide range of domains such as medicine, mathematics, and the arts, allowing us to capture a diverse set of modalities and task combinations. 
For each domain, we systematically identify the core perception and reasoning skills required to answer the associated questions, ensuring that our expert pool is both modular and generalizable across tasks. This design allows our framework to scale beyond any single dataset and adapt to new domains with minimal effort.

In Fig.~\ref{fig:MEXA architecture}, we illustrate the complete set of expert modules assembled in our framework. Specifically, we categorize these experts into four distinct types: (1) \textbf{Perceptual Experts}, specialized in extracting visual information directly from images, videos, audio, 3D point clouds and medical images; (2) \textbf{Text Extraction Experts}, which leverage optical character recognition (OCR) to distill textual content from visuals such as slides or embedded text; (3) \textbf{Structured Experts}, designed to analyze structured visual data such as charts, tables, and diagrams; and (4) \textbf{Mathematical Reasoning Expert}, 
focused on interpreting and solving questions related to mathematical and geometric equations presented in LaTeX format.
Furthermore, for each expert category, we design skill-specialized experts that target distinct reasoning demands across modalities and task types. We focus on aligning each expert’s functionality with the specific requirements of the task.
For instance, for 2D image perception, we design experts aligned with skills corresponding to different levels of reasoning granularity: a fine-grained image description expert, responsible for generating comprehensive captions covering multiple visual elements and their relationships, and a concise summarization expert, which produces brief, high-level summaries highlighting only the salient visual content.
Within the domain of 3D scene understanding, we develop two experts with specialized skills for different task demands: a general 3D scene expert, which provides descriptions capturing the overall spatial layout and major objects in the scene, and a situated 3D expert, designed to generate detailed, viewpoint-grounded descriptions reflecting the specific perspective of an embodied agent positioned within the environment.

\paragraph{Design Principle 2: Unified Textual Representation.} 
We design expert models that convert diverse modality-specific inputs into a shared textual representation, facilitating the integration of heterogeneous multimodal data. 
To achieve this, we leverage a series of state-of-the-art captioning models combined with modality- and skill-specific prompting strategies.
Each expert in our framework functions as a captioner (descriptor), converting modality- and skill-specific information into a natural language format. 
This textual abstraction further enables interpretable perception and supports reasoning at varying levels of complexity. 
Formally, given a set of supported input modalities \( \mathcal{M}\), the objective is to obtain a unified textual representation \( x \) by converting the relevant modality-specific input into natural language. 
This is accomplished through a set of modality-specific expert functions $E_i$, and textual output from each expert is denoted as:
\begin{equation}
    x_i = E_i(m_i, p_i),
\end{equation}
where $p_i$ is a carefully designed prompt that reflects the reasoning objective for different modalities and reasoning demands.

\begin{figure*}
    \centering
    \includegraphics[width=0.95\linewidth]{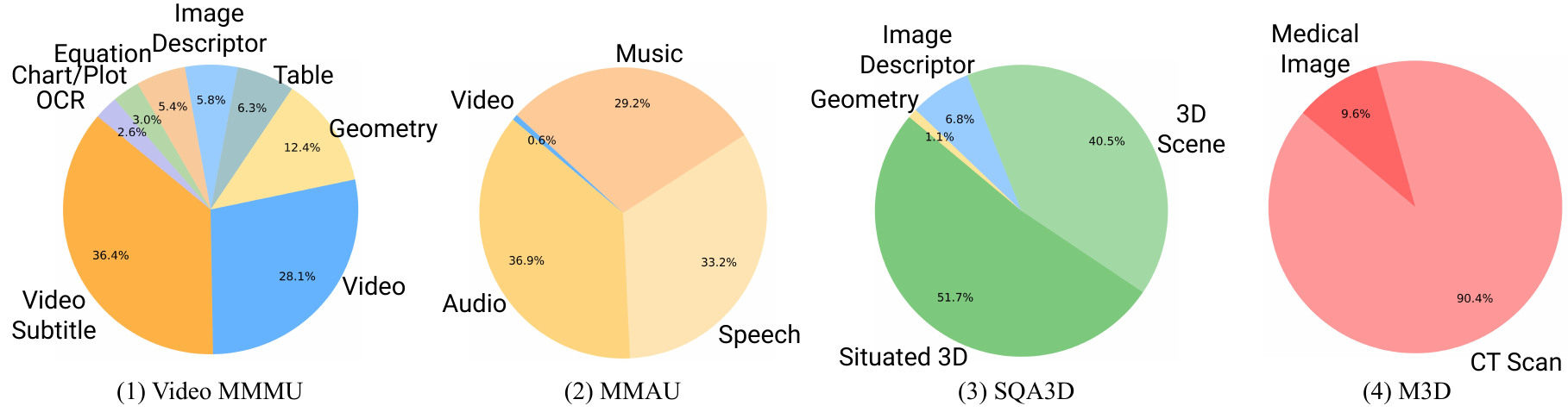}
    \caption{Expert distributions selected by \model{} across different benchmarks, covering video (Video-MMMU), audio (MMAU), 3D (SQA3D), and medical imaging (M3D).}
    \label{fig:skill distribution}
    \vspace{-3mm}
\end{figure*}
\subsubsection{Expert Selection Module}   
\label{MLLM Router}
After constructing the pool of expert models, the next step is to determine which experts to activate for a given task. We design an expert selection module that leverages a multimodal LLM (MLLM) as a router for different experts. 
This expert selection process leverages the commonsense reasoning capabilities of the MLLM, allowing it to deeply understand the semantics of the task, the relationships among modalities, and the contextual intent behind the question when deciding which experts to activate. 
As illustrated in Fig.~\ref{fig:MEXA architecture}, the router takes task context and question as input. Task context provides high-level guidance on the task type, and helps the router infer the set of skills required to answer the question and explicitly constrains the router by ensuring that only relevant experts are considered.
Based on these inputs, the expert selection module activates a subset of expert models most suited to addressing the question.

Formally, given question $Q_\text{ques}$, a task description $T_r$, also a set of available modality inputs $\mathcal{M}$, the MLLM router selects a subset of expert models based on needed modality and tasks. We denote the selected experts models as \(\{E_s\}_{s \in \mathcal{S}}\), where $\mathcal{S} \subseteq \mathcal{M}$. Once the experts are selected, each expert \(\{E_s\}\) receives the specific prompt $p_s$ and transforms the input into a natural language description $x_s$. The resulting set of expert-generated textual output is represented as :

\begin{equation}
\begin{split}
\mathcal{T} = \{x_s = &E_s(m_s,p_s)\ |\ 
E_s \in \texttt{Router}(\\
&Q_{\text{ques}}, T_r, \mathcal{M}), 
s \in \mathcal{S} \subseteq \mathcal{M} \}\\
\end{split}
\end{equation}
where $\mathcal{T}$ represents the set of intermediate textual outputs generated by the selected experts, serving as a unified and interpretable representation that is later aggregated to produce the final answer.

\subsubsection{Aggregating Expert Information via Reasoning}
\label{aggregator}
Once the selected experts have generated modality- and task-specific textual representations, we employ a reasoning module based on LRM, selected for its strong long-context reasoning and deep inference capabilities. This LRM-based aggregator systematically integrates the diverse outputs from the selected experts and reasons over them to generate the final answer. 
Formally, the expert outputs are represented as $\mathcal{T} = \{t_1, t_2, \dots, t_k\}$, where $k$ denotes the number of selected expert.
The aggregator inputs the entire set $\mathcal{T}$, along with the task description for the expert, which is denoted as $T_a$, and generates the final answer, denoted as: 

\begin{equation}
     \begin{split}
     A = \texttt{Aggregator}(\mathcal{T}; T_a)
     \end{split}
\end{equation}

By aggregating expert-generated textual outputs and performing reasoning over these unified representations, \model{} enables more accurate, interpretable, and task-aligned answers,  demonstrating clear advantages over single end-to-end models that process implicit representations without explicit decomposition or modular coordination.

\section{Experimental Results}
\subsection{Experimental Setup}
We use GPT-4o~\cite{openai2024gpt4ocard} as our multimodal expert selection module to dynamically select the most relevant experts based on the input modalities and task requirements. Subsequently, we employ Deepseek as our aggregator to reason over the expert-generated textual information to obtain the final answer. 

\begin{table*}
\renewcommand{\arraystretch}{1.2}
\small
\centering

\label{tab:trackresult}
\setlength{\tabcolsep}{1.mm}
\resizebox{1\textwidth}{!}{
\begin{tabular}{@{}l @{\hspace{1pt}}c c @{\hspace{3pt}}c @{\hspace{3pt}}c !{\vrule width 0.5pt} c c c c c c@{}}
\toprule
\textbf{Method} & \textbf{Overall} & \multicolumn{3}{c}{\textbf{Results by Track}} & \multicolumn{6}{c}{\textbf{ Results by Discipline}} \\
\cmidrule(lr){3-5} \cmidrule(lr){6-11}
&&\textbf{Perception} & \textbf{Comprehension} & \textbf{Adaptation}& \textbf{Art.} & \textbf{Biz.} & \textbf{Sci.}& \textbf{Med.} & \textbf{Hum.} & \textbf{Eng.} \\
\midrule
Random Choice & $14.0$ & $12.0$ & $14.0$ & $16.0$ & $11.1$ & $12.9$ & $12.1$ & $22.5$ & $10.5$ & $13.6$ \\
\midrule
\multicolumn{11}{l}{\textbf{Open-source LMMs}} \\
LLaVA-OneVision-72B \cite{li2024llava}  & $48.3$ & $59.7$ & $42.3$ & $43.0$ & $61.9$ & $46.21$ & $40.2$ & $54.3$ & $60.0$ & $44.0$\\
LLaVA-Video-72B \cite{zhang2024video} & $49.7$ & $59.7$ & $46.0$ & $43.3$ & $69.8$ & $44.7$ & $41.7$ & $58.9$ & $57.1$ & $45.1$ \\
Aria~\cite{li2024aria} (8 $\times$ 3.5B) & $50.8$ & $65.7$ & $46.7$ & $40.0$ & ${71.4}$ & $47.7$ & $44.7$ & $58.9$ & $62.9$ & $43.7$  \\ 
\midrule
 \multicolumn{11}{l}{\textbf{Proprietary LMMs}} \\
Gemini 1.5 Flash \cite{geminiteam2024gemini15unlockingmultimodal}  & $49.8$ & $57.3$ & $49.0$ & $43.0$ & $63.5$ & $53.0$ & $43.2$ & $49.6$ & $59.1$ & $45.7$ \\
Gemini 1.5 Pro \cite{geminiteam2024gemini15unlockingmultimodal}  & $53.9$ & $59.0$ & $53.3$ & $49.3$ & $57.1$ & $59.1$ & $49.1$ & $57.4$ & $58.1$ & $50.3$ \\ 
GPT-4o \cite{openai2024gpt4ocard} & $61.2$ & $66.0$  & $62.0$ & ${55.7}$ & $69.5$ & $66.9$ & $51.6$ & ${64.8}$ & $69.5$ & $57.1$ \\
Claude-3.5-Sonnet \cite{anthropic_claude35_sonnet_2024} & $65.8$ & ${72.0}$ & ${69.7}$ & ${55.7}$ & $66.7$ & ${75.0}$ & ${56.1}$ & $58.1$ & ${75.2}$ & ${66.1}$ \\ \midrule
\textcolor{gray}{\textit{Human Expert}} & \textcolor{gray}{$74.4$} & \textcolor{gray}{$84.3$} & \textcolor{gray}{$78.7$} & \textcolor{gray}{$60.3$} & \textcolor{gray}{$81.0$} & \textcolor{gray}{$78.8$} & \textcolor{gray}{$74.2$} & \textcolor{gray}{$70.5$} & \textcolor{gray}{$84.8$} & \textcolor{gray}{$69.9$} \\
\midrule
\textbf{MEXA (Ours)} & $\mathbf{71.5}$ & $\mathbf{77.0}$ & $\mathbf{76.7}$ & $\mathbf{60.0}$ & $\mathbf{76.2}$ & $\mathbf{77.0}$ & $\mathbf{63.8}$ & $\mathbf{75.8}$ & $\mathbf{78.1}$ & $\mathbf{67.7}$ \\ 

\bottomrule
\end{tabular}
}
\caption{
Video-MMMU Evaluation Results across three cognitive tracks (Perception, Comprehension, Adaptation) and six disciplines (Art, Business, Science, Medicine, Humanities, Engineering). We highlight the best model performance for each metric.
}
\end{table*}

To effectively extract textual information from diverse expert modules, we adopt the widely used captioners to generate high-quality textual descriptions across modalities. Specifically, for perceptual experts, we employ different captioning approaches to each modality. For 2D image inputs, we utilize OmniCaptioner-Qwen2-5-7B~\cite{lu2025omnicaptionercaptionerrule} to generate both detailed and concise descriptions. For video modality experts, we adopt NVILA-8B~\cite{liu2024nvila} to generate comprehensive video-level captions.
For experts specializing in 3D scene understanding, we integrate LEO-Vicuna7B~\cite{huang2024embodied} to obtain general, top-down scene descriptions and Spartun3D-Vicuna7B~\cite{zhang2024spartun3d} for generating situated captions explicitly grounded in the agent’s viewpoint. Regarding audio modality experts, we employ Qwen-2.5-Omni~\cite{xu2025qwen2}, which is adept at generating contextualized captions for both speech and music. For all non-perceptual experts, we apply  OmniCaptioner-qwen-5-7B~\cite{lu2025omnicaptionercaptionerrule} with prompts designed to emphasize the specific functionality of each expert. We provide detailed prompts for the selection module, aggregator, and all expert models in the Appendix.

\subsection{Evaluation Datasets}
We validate our framework across various challenging multimodal tasks, including Video Reasoning (Video-MMMU \citep{hu2025videommmuevaluatingknowledgeacquisition}), Audio QA (MMAU \citep{sakshi2024mmaumassivemultitaskaudio}), 3D Situated Reasoning (SQA3D \citep{ma2023sqa3dsituatedquestionanswering}), and Medical QA (M3D \citep{bai2024m3d}). We specifically chose these benchmarks because they represent diverse reasoning complexities, modality interactions, and practical application scenarios. Please see more details in the Appendix.

\subsection{Quantitative Results}

We evaluate \model{} on all datasets under the multiple-choice QA setting, and report performance based on standard accuracy metrics across all experiments.

\noindent\textbf{Video-base Multimodal Reasoning.} 
For the video-based multimodal reasoning capability, we evaluate MXEA on the Video-MMMU benchmark.
We compare our methods with three types of baselines: open-source large multimodal models (LMMs) \cite{li2024llava, li2024aria}, proprietary LMMs \cite{openai2024gpt4ocard, anthropic_claude35_sonnet_2024, geminiteam2024gemini15unlockingmultimodal}, and human experts \cite{hu2025videommmuevaluatingknowledgeacquisition}. 
Results show that \model{} significantly outperforms the leading open-source LLM \cite{li2024aria} by $23.6\%$ in overall accuracy. 
Moreover, our method outperforms the powerful MLLM like GPT-4o \cite{openai2024gpt4ocard} by $6\%$ in overall accuracy. 
Our method also achieves performance on par with the human expert on most metrics, showing the effectiveness of the proposed framework. 
Specifically, we see strong results across all six disciplines that validate the effectiveness of the combination of expert modules.
For comparison, GPT-4o \cite{openai2024gpt4ocard} achieves only $51.6\%$ and $57.1\%$ on science and engineering, performing $12.2\%$ and $10.5\%$ lower than \model{}, respectively. This gap highlights the advantage of explicitly aggregating and reasoning over information from multiple specialized experts, rather than relying solely on a single multimodal LLM.

\begin{table*}
\centering
\begin{minipage}[t]{0.41\textwidth}
\small
\centering
\setlength{\tabcolsep}{1.mm}
\resizebox{1\textwidth}{!}{
\begin{tabular}{lcccc}
\toprule
\textbf{Method} & \textbf{Sound} & \textbf{Music} & \textbf{Speech} & \textbf{Average} \\ \midrule
\textcolor{gray}{GPT-4o~\cite{openai2024gpt4ocard} w/ caption} & \textcolor{gray}{$39.3$} & \textcolor{gray}{$39.5$} & \textcolor{gray}{$58.3$} & \textcolor{gray}{$45.7$} \\
\midrule
GAMA-7B~\cite{ghosh2024gama} & $41.4$ & $32.3$ & $18.9$ & $30.9$\\
MULLaMA-7B~\cite{liu2024music}&$40.8$& $32.6$&$22.2$&$31.9$\\
SALAMONN-13B~\cite{tang2023salmonn} & $41.0$ & $34.8$ & $25.5$ & $33.7$\\

\textbf{MEXA (Ours)}  &$\mathbf{45.1}$& $\mathbf{40.7}$ & $\mathbf{51.8}$ & $\mathbf{45.9}$\\ 
\bottomrule
\end{tabular}
}
\caption{MMAU evaluation results across three different audio question types.}
\label{tab:audio_main}
\end{minipage}
\hfill\begin{minipage}[t]{0.57\textwidth}
\small
\setlength{\tabcolsep}{1.mm}
\resizebox{\textwidth}{!}{
\begin{tabular}{lccccccr}
\toprule
\textbf{Method} & \textbf{What} & \textbf{Is} & \textbf{How} & \textbf{Can} & \textbf{Which} & \textbf{Others} &  \textbf{Avg.} \\ \midrule
CREMA~\cite{yu2024crema} &$23.2$&$\mathbf{52.5}$&$34.8$&$41.5$&$34.1$& $37.5$ & $37.3$\\
LEO~\cite{huang2024embodied} & $10.2$ & $15.7$ & $11.6$ & $9.4$ & $10.6$ & $16.9$ &  $12.4$ \\
Spartun3D~\cite{zhang2024spartun3d} & $22.3$ & $50.9$ & $33.8$ & $40.9$ & $34.4$ & $34.5$ & $36.1$ \\
\textbf{MEXA (Ours)} & $\mathbf{23.4}$ & $51.1$ & $\mathbf{35.2}$ & $\mathbf{42.5}$ & $\mathbf{36.9}$ & $\mathbf{37.9}$ & $\mathbf{37.8}$\\
\bottomrule
\end{tabular}
}
\caption{SQA3D evaluation results across fine-grained question types.}\vspace{-0.05in}
\label{tab:sqa3d_main}
\end{minipage}
\end{table*}







\begin{table*}[t]
\centering
\begin{minipage}[t]{0.57\textwidth}
\vspace{-0.05in}
\small
\setlength{\tabcolsep}{1.mm}
\resizebox{\textwidth}{!}{
\begin{tabular}{lcccccr}
\toprule
\textbf{Method} & \textbf{Plane} & \textbf{Phase} & \textbf{Organ} & \textbf{Abnormality} & \textbf{Location} & \textbf{Avg.} \\ \midrule
CREMA~\cite{yu2024crema} & $14.9$ & $26.7$ & $15.9$ & $17.3$ & $13.0$ & $17.2$ \\
MiniCPM-o~\cite{yao2024minicpm} & 60.7 & 39.2 & 35.2 &  \textbf{53.0} & 42.0 & 44.7 \\
GPT-4o & \textbf{83.3} & 42.7 & 50.0 & 41.3& 41.3 & 51.7 \\
\textbf{MEXA (Ours)} & ${65.0}$ & $\textbf{48.1}$ & $\textbf{60.9}$ & $44.8$ & $\textbf{48.0}$ & $\textbf{53.3}$ \\
\bottomrule
\end{tabular}
}
\caption{M3D evaluation results across different kinds of medical QA types.
\label{tab:m3d_main}
}
\end{minipage}
\hfill
\begin{minipage}[t]{0.4\textwidth}
\vspace{-0.05in}
\small
\setlength{\tabcolsep}{1.mm}
\resizebox{\textwidth}{!}{
\begin{tabular}{cccc}
\toprule
\textbf{Router} & \textbf{Aggregator} & \textbf{Video-MMMU} & \textbf{M3D} \\ \midrule
Qwen2.5-VL (7B) & GPT-4o & $57.4$ & $34.7$ \\
Qwen2.5-VL (7B) & DeepSeek & $70.4$ & $45.8$ \\
GPT-4o & GPT-4o & $58.9$ & $48.2$ \\
GPT-4o & DeepSeek & $\mathbf{71.5}$ & $\mathbf{53.4}$ \\
\bottomrule
\end{tabular}
}
\caption{Ablation study on MLLM router and LLM aggregator.}
\label{tab:ablation_router}
\end{minipage}
\end{table*}

\noindent\textbf{Audio-based Multimodal Reasoning.} 
\cref{tab:audio_main} presents the results of the MMAU benchmark, evaluating models on three audio QA types: \textit{Sound}, \textit{Music}, and \textit{Speech}. Our proposed \model{} consistently outperforms all audio large language models, including GAMA, MuLLaMA, and SALAMONN, across all categories. 
Notably, \model{} achieves a substantial improvement of $4.1\%$, $5.9\%$, and $26.3\%$ over the best-performing Audio LLM baselines, demonstrating its strong and generalizable ability to understand diverse non-visual sensory inputs, such as audio. In contrast, existing models struggle particularly with complex music and speech QA tasks. 
Compared to GPT-4o with generated captions using~\cite{kim2024enclap,liu2024music,radford2023robust}, \model{} achieves comparable overall performance while maintaining a more balanced accuracy across all audio types. These results validate the effectiveness of our modular expert-based approach in addressing the diverse challenges of audio understanding.

\noindent\textbf{3D Scene Understanding.} 
Results on the SQA3D benchmark shown in \cref{tab:sqa3d_main} demonstrate the effectiveness of our method in 3D situated reasoning tasks. By integrating situated captions and general scene descriptions through our routing and aggregation system, \model{} consistently improves accuracy across all question types. Specifically, our approach achieves a $2\%$ accuracy gain over SOTA 3D-based LLMs, which rely on general-purpose encoder-fusion strategies with static unified architectures. This improvement highlights the advantage of our dynamic, modality-aware expert selection and aggregation framework in the 3D domain.

\noindent\textbf{Medical QA.}
To demonstrate the effectiveness of our approach beyond general domains such as audio and video, we additionally evaluate \model{} on the medical video QA domain using the M3D benchmark. This benchmark includes five distinct medical QA types:  Plane classification, Phase recognition, Organ identification, Abnormality detection, and Location estimation. As shown in~\cref{tab:m3d_main}, \model{} significantly outperforms strong general-purpose MLLM, CREMA, MiniCPM-o, and GPT-4o, which directly process 3D scan images as input, and observe that \model{} consistently achieves superior performance across most categories, improving average accuracy by 36.1\%, 8.7\% and 1.6\%, respectively. These results underscore \model{}'s robust capability in medical content understanding, even without extensive domain-specific fine-tuning.

\subsection{Ablation Study and Analysis}

\paragraph{Ablation on Router and Aggregator.} Table~\ref{tab:ablation_router} presents ablation results for different router and aggregator configurations. We observe that, in addition to the reasoning strength of the model itself, the correct deployment and targeted usage of these models significantly contribute to performance. Specifically, we compare the effectiveness of Qwen2.5-VL and GPT-4o as multimodal expert selection modules, alongside GPT-4o and DeepSeek as aggregators, across two benchmarks: Video-MMMU and M3D. We observe that GPT-4o consistently outperforms Qwen2.5-VL when utilized as the router, indicating its superior capability in multimodal expert selection.
Furthermore, DeepSeek demonstrates better performance than GPT-4o in the aggregator role, underscoring its effectiveness in reasoning over the outputs from specialized experts. This superior performance indicates that our aggregator requires strong reasoning abilities, including the effective handling of long-context information and nuanced inference across diverse expert outputs, in which DeepSeek particularly excels.

\paragraph{Expert Distributions.} 
In \cref{fig:skill distribution}, we visualize the expert selection distributions from \model{}'s expert selection module. For the Video-MMMU benchmark, multiple experts are frequently selected, aligning well with the benchmark's inherent multimodal and multidisciplinary nature, which demands diverse skills for comprehensive video understanding. In the MMAU audio reasoning benchmark, the audio-specific experts~(music, audio, and speech) are consistently activated, with their selections evenly distributed, reflecting balanced reliance on all audio modalities. For the 3D reasoning task (SQA3D), the situated 3D and general 3D scene experts are predominantly selected, with occasional inclusion of the image descriptor expert. Finally, for the medical dataset (M3D-VQA), the CT scan expert is primarily selected, complemented occasionally by the medical image expert. These observations confirm that our selection module effectively and adaptively matches experts to the specific modality requirements and reasoning contexts of each task. We provide qualitative examples of Video-MMMU and SQA3D in the Appendix.

\section{Conclusion}
In this paper, we introduce \model{}, a dynamic multimodal reasoning framework that adaptively leverages a diverse set of expert modules with the skills for specific modalities and tasks. By effectively integrating an expert selection module and aggregation mechanism, our method achieves superior performance across a variety of challenging benchmarks, including Video Reasoning, Audio QA, 3D Situated Reasoning, and Medical QA. Extensive experimental analyses and visualization results demonstrate the clear advantages of our dynamic and modality-aware design, highlighting the significance of selectively combining complementary expert skills. We believe that our framework provides a robust foundation for future research in generalized multimodal reasoning, enabling models to tackle increasingly complex real-world multimodal applications.




\section*{Limitations}
Our framework achieves competitive reasoning across diverse domains by leveraging modular expert outputs. However, since it relies on frozen pre-trained experts, its reasoning fidelity can be constrained by their inherent capabilities, occasionally leading to incorrect or hallucinated rationales. Furthermore, the quality of the final prediction is affected by the precision and expressiveness of individual experts. As the ecosystem of multimodal and language models evolves, integrating more advanced experts holds promise for improving both accuracy and robustness without altering the overall framework. 

\section{Acknowledgment}
We thank the reviewers and area chairs for their helpful feedback. This work was supported by the National Institutes of Health (NIH) under other transactions 1OT2OD038045-01, ARO Award W911NF2110220, ONR Grant N00014-23-1-2356, DARPA ECOLE Program No. HR00112390060, NSF-AI Engage Institute DRL211263, and a Capital One Research Award. The views, opinions, and/or findings contained in this article are those of the authors and not of the funding agency.
{
\bibliography{yu}

\begin{thebibliography}{81}
\providecommand{\natexlab}[1]{#1}

\bibitem[{{Anthropic}(2024)}]{anthropic_claude35_sonnet_2024}
{Anthropic}. 2024.
\newblock \href {https://www.anthropic.com/news/claude-3-5-sonnet} {Claude 3.5 sonnet}.

\bibitem[{Antol et~al.(2015)Antol, Agrawal, Lu, Mitchell, Batra, Zitnick, and Parikh}]{antol2015vqa}
Stanislaw Antol, Aishwarya Agrawal, Jiasen Lu, Margaret Mitchell, Dhruv Batra, C~Lawrence Zitnick, and Devi Parikh. 2015.
\newblock Vqa: Visual question answering.
\newblock In \emph{Proceedings of the International Conference on Computer Vision (ICCV)}.

\bibitem[{Azam et~al.(2022)Azam, Khan, Salahuddin, Rehman, Khan, Khan, Kadry, and Gandomi}]{azam2022review}
Muhammad~Adeel Azam, Khan~Bahadar Khan, Sana Salahuddin, Eid Rehman, Sajid~Ali Khan, Muhammad~Attique Khan, Seifedine Kadry, and Amir~H Gandomi. 2022.
\newblock A review on multimodal medical image fusion: Compendious analysis of medical modalities, multimodal databases, fusion techniques and quality metrics.
\newblock \emph{Computers in biology and medicine}, 144:105253.

\bibitem[{Bai et~al.(2024)Bai, Du, Huang, Meng, and Zhao}]{bai2024m3d}
Fan Bai, Yuxin Du, Tiejun Huang, Max Q-H Meng, and Bo~Zhao. 2024.
\newblock M3d: Advancing 3d medical image analysis with multi-modal large language models.
\newblock \emph{arXiv preprint arXiv:2404.00578}.

\bibitem[{Brohan et~al.(2023)Brohan, Brown, Carbajal, Chebotar, Chen, Choromanski, Ding, Driess, Dubey, Finn et~al.}]{brohan2023rt}
Anthony Brohan, Noah Brown, Justice Carbajal, Yevgen Chebotar, Xi~Chen, Krzysztof Choromanski, Tianli Ding, Danny Driess, Avinava Dubey, Chelsea Finn, and 1 others. 2023.
\newblock Rt-2: Vision-language-action models transfer web knowledge to robotic control.
\newblock \emph{arXiv preprint arXiv:2307.15818}.

\bibitem[{Cai et~al.(2023)Cai, Fang, and Li}]{cai2023pre}
Linqin Cai, Haodu Fang, and Zhiqing Li. 2023.
\newblock Pre-trained multilevel fuse network based on vision-conditioned reasoning and bilinear attentions for medical image visual question answering.
\newblock \emph{The Journal of Supercomputing}, 79(12):13696--13723.

\bibitem[{Cao et~al.(2025)Cao, Zhang, Li, Li, Joty, and Carenini}]{cao2025multi2}
Juntai Cao, Xiang Zhang, Raymond Li, Chuyuan Li, Shafiq Joty, and Giuseppe Carenini. 2025.
\newblock Multi2: Multi-agent test-time scalable framework for multi-document processing.
\newblock \emph{arXiv preprint arXiv:2502.20592}.

\bibitem[{Chen et~al.(2025)Chen, Yun, Stengel-Eskin, Chen, and Bansal}]{chen2025symbolic}
Justin Chih-Yao Chen, Sukwon Yun, Elias Stengel-Eskin, Tianlong Chen, and Mohit Bansal. 2025.
\newblock Symbolic mixture-of-experts: Adaptive skill-based routing for heterogeneous reasoning.
\newblock \emph{arXiv preprint arXiv:2503.05641}.

\bibitem[{Chen et~al.(2023{\natexlab{a}})Chen, Li, Wang, Zhao, Sun, Zhu, and Liu}]{chen2024vast}
Sihan Chen, Handong Li, Qunbo Wang, Zijia Zhao, Mingzhen Sun, Xinxin Zhu, and Jing Liu. 2023{\natexlab{a}}.
\newblock Vast: A vision-audio-subtitle-text omni-modality foundation model and dataset.
\newblock In \emph{Advances in Neural Information Processing Systems (NeurIPS)}.

\bibitem[{Chen et~al.(2023{\natexlab{b}})Chen, Shen, Ding, Chen, Zhao, Learned-Miller, and Gan}]{chen2023mod}
Zitian Chen, Yikang Shen, Mingyu Ding, Zhenfang Chen, Hengshuang Zhao, Erik~G Learned-Miller, and Chuang Gan. 2023{\natexlab{b}}.
\newblock Mod-squad: Designing mixtures of experts as modular multi-task learners.
\newblock In \emph{Proceedings of the IEEE/CVF Conference on Computer Vision and Pattern Recognition}, pages 11828--11837.

\bibitem[{Chowdhury et~al.(2023)Chowdhury, Zhang, Wang, Liu, and Chen}]{chowdhury2023patch}
Mohammed Nowaz~Rabbani Chowdhury, Shuai Zhang, Meng Wang, Sijia Liu, and Pin-Yu Chen. 2023.
\newblock Patch-level routing in mixture-of-experts is provably sample-efficient for convolutional neural networks.
\newblock In \emph{International Conference on Machine Learning}, pages 6074--6114. PMLR.

\bibitem[{Dai et~al.(2022)Dai, Dong, Ma, Zheng, Sui, Chang, and Wei}]{dai2022stablemoe}
Damai Dai, Li~Dong, Shuming Ma, Bo~Zheng, Zhifang Sui, Baobao Chang, and Furu Wei. 2022.
\newblock Stablemoe: Stable routing strategy for mixture of experts.
\newblock In \emph{Proceedings of the 60th Annual Meeting of the Association for Computational Linguistics (Volume 1: Long Papers)}, pages 7085--7095.

\bibitem[{Driess et~al.(2023)Driess, Xia, Sajjadi, Lynch, Chowdhery, Wahid, Tompson, Vuong, Yu, Huang et~al.}]{driess2023palm}
Danny Driess, Fei Xia, Mehdi~SM Sajjadi, Corey Lynch, Aakanksha Chowdhery, Ayzaan Wahid, Jonathan Tompson, Quan Vuong, Tianhe Yu, Wenlong Huang, and 1 others. 2023.
\newblock Palm-e: An embodied multimodal language model.

\bibitem[{Durante et~al.(2024)Durante, Huang, Wake, Gong, Park, Sarkar, Taori, Noda, Terzopoulos, Choi et~al.}]{durante2024agent}
Zane Durante, Qiuyuan Huang, Naoki Wake, Ran Gong, Jae~Sung Park, Bidipta Sarkar, Rohan Taori, Yusuke Noda, Demetri Terzopoulos, Yejin Choi, and 1 others. 2024.
\newblock Agent ai: Surveying the horizons of multimodal interaction.
\newblock \emph{arXiv preprint arXiv:2401.03568}.

\bibitem[{Ghosh et~al.(2024)Ghosh, Kumar, Seth, Evuru, Tyagi, Sakshi, Nieto, Duraiswami, and Manocha}]{ghosh2024gama}
Sreyan Ghosh, Sonal Kumar, Ashish Seth, Chandra Kiran~Reddy Evuru, Utkarsh Tyagi, S~Sakshi, Oriol Nieto, Ramani Duraiswami, and Dinesh Manocha. 2024.
\newblock Gama: A large audio-language model with advanced audio understanding and complex reasoning abilities.
\newblock \emph{arXiv preprint arXiv:2406.11768}.

\bibitem[{Girdhar et~al.(2023)Girdhar, El-Nouby, Liu, Singh, Alwala, Joulin, and Misra}]{girdhar2023imagebind}
Rohit Girdhar, Alaaeldin El-Nouby, Zhuang Liu, Mannat Singh, Kalyan~Vasudev Alwala, Armand Joulin, and Ishan Misra. 2023.
\newblock Imagebind: One embedding space to bind them all.
\newblock In \emph{Proceedings of the IEEE International Conference on Computer Vision and Pattern Recognition (CVPR)}.

\bibitem[{Gong et~al.(2023)Gong, Lyu, Zhang, Wang, Zheng, Zhao, Liu, Zhang, Luo, and Chen}]{gong2023multimodalgpt}
Tao Gong, Chengqi Lyu, Shilong Zhang, Yudong Wang, Miao Zheng, Qian Zhao, Kuikun Liu, Wenwei Zhang, Ping Luo, and Kai Chen. 2023.
\newblock Multimodal-gpt: A vision and language model for dialogue with humans.
\newblock \emph{2305.04790}.

\bibitem[{Guo et~al.(2025)Guo, Song, Zhang, Liu, and Liu}]{guo2025rethinking}
Xiao Guo, Xiufeng Song, Yue Zhang, Xiaohong Liu, and Xiaoming Liu. 2025.
\newblock Rethinking vision-language model in face forensics: Multi-modal interpretable forged face detector.
\newblock In \emph{Proceedings of the Computer Vision and Pattern Recognition Conference}, pages 105--116.

\bibitem[{Han et~al.(2023)Han, Gong, Zhang, Wang, Zhang, Lin, Qiao, Gao, and Yue}]{han2023onellm}
Jiaming Han, Kaixiong Gong, Yiyuan Zhang, Jiaqi Wang, Kaipeng Zhang, Dahua Lin, Yu~Qiao, Peng Gao, and Xiangyu Yue. 2023.
\newblock Onellm: One framework to align all modalities with language.
\newblock \emph{arXiv preprint arXiv:2312.03700}.

\bibitem[{He et~al.(2020)He, Zhang, Mou, Xing, and Xie}]{he2020pathvqa}
Xuehai He, Yichen Zhang, Luntian Mou, Eric Xing, and Pengtao Xie. 2020.
\newblock Pathvqa: 30000+ questions for medical visual question answering.
\newblock \emph{arXiv preprint arXiv:2003.10286}.

\bibitem[{Hong et~al.(2023)Hong, Zhen, Chen, Zheng, Du, Chen, and Gan}]{hong20233d}
Yining Hong, Haoyu Zhen, Peihao Chen, Shuhong Zheng, Yilun Du, Zhenfang Chen, and Chuang Gan. 2023.
\newblock 3d-llm: Injecting the 3d world into large language models.
\newblock \emph{Advances in Neural Information Processing Systems}, 36:20482--20494.

\bibitem[{Hou et~al.(2023)Hou, Dai, He, Dai, and Nie{\ss}ner}]{hou2023mask3d}
Ji~Hou, Xiaoliang Dai, Zijian He, Angela Dai, and Matthias Nie{\ss}ner. 2023.
\newblock Mask3d: Pre-training 2d vision transformers by learning masked 3d priors.
\newblock In \emph{Proceedings of the IEEE International Conference on Computer Vision and Pattern Recognition (CVPR)}.

\bibitem[{Hou et~al.(2021)Hou, Xie, Graham, Dai, and Nie{\ss}ner}]{hou2021pri3d}
Ji~Hou, Saining Xie, Benjamin Graham, Angela Dai, and Matthias Nie{\ss}ner. 2021.
\newblock Pri3d: Can 3d priors help 2d representation learning?
\newblock In \emph{Proceedings of the International Conference on Computer Vision (ICCV)}.

\bibitem[{Hu et~al.(2025{\natexlab{a}})Hu, Wu, Pu, Xiao, Zhang, Yue, Li, and Liu}]{hu2025video}
Kairui Hu, Penghao Wu, Fanyi Pu, Wang Xiao, Yuanhan Zhang, Xiang Yue, Bo~Li, and Ziwei Liu. 2025{\natexlab{a}}.
\newblock Video-mmmu: Evaluating knowledge acquisition from multi-discipline professional videos.
\newblock \emph{arXiv preprint arXiv:2501.13826}.

\bibitem[{Hu et~al.(2025{\natexlab{b}})Hu, Wu, Pu, Xiao, Zhang, Yue, Li, and Liu}]{hu2025videommmuevaluatingknowledgeacquisition}
Kairui Hu, Penghao Wu, Fanyi Pu, Wang Xiao, Yuanhan Zhang, Xiang Yue, Bo~Li, and Ziwei Liu. 2025{\natexlab{b}}.
\newblock \href {https://arxiv.org/abs/2501.13826} {Video-mmmu: Evaluating knowledge acquisition from multi-discipline professional videos}.
\newblock \emph{Preprint}, arXiv:2501.13826.

\bibitem[{Huang et~al.(2024)Huang, Yong, Ma, Linghu, Li, Wang, Li, Zhu, Jia, and Huang}]{huang2024embodied}
Jiangyong Huang, Silong Yong, Xiaojian Ma, Xiongkun Linghu, Puhao Li, Yan Wang, Qing Li, Song-Chun Zhu, Baoxiong Jia, and Siyuan Huang. 2024.
\newblock An embodied generalist agent in 3d world.
\newblock In \emph{Proceedings of the 41st International Conference on Machine Learning}, pages 20413--20451.

\bibitem[{Huang et~al.(2023)Huang, Gu, Hou, Wu, Wang, Yu, and Han}]{huang2022large}
Jiaxin Huang, Shixiang~Shane Gu, Le~Hou, Yuexin Wu, Xuezhi Wang, Hongkun Yu, and Jiawei Han. 2023.
\newblock Large language models can self-improve.
\newblock In \emph{Proceedings of the Conference on Empirical Methods in Natural Language Processing (EMNLP)}.

\bibitem[{Hudson and Manning(2019)}]{hudson2019gqa}
Drew~A Hudson and Christopher~D Manning. 2019.
\newblock Gqa: A new dataset for real-world visual reasoning and compositional question answering.
\newblock In \emph{Proceedings of the IEEE International Conference on Computer Vision and Pattern Recognition (CVPR)}.

\bibitem[{Hurst et~al.(2024)Hurst, Lerer, Goucher, Perelman, Ramesh, Clark, Ostrow, Welihinda, Hayes, Radford, Mądry, Baker-Whitcomb, Beutel, Borzunov, Carney, Chow, Kirillov, Nichol, Paino, Renzin, Passos, Kirillov, Christakis, Conneau, Kamali, Jabri, Moyer, Tam, Crookes, Tootoochian, Tootoonchian, Kumar, Vallone, Karpathy, Braunstein, Cann, Codispoti, Galu, Kondrich, Tulloch, Mishchenko, Baek, Jiang, Pelisse, Woodford, Gosalia, Dhar, Pantuliano, Nayak, Oliver, Zoph, Ghorbani, Leimberger, Rossen, Sokolowsky, Wang, Zweig, Hoover, Samic, McGrew, Spero, Giertler, Cheng, Lightcap, Walkin, Quinn, Guarraci, Hsu, Kellogg, Eastman, Lugaresi, Wainwright, Bassin, Hudson, Chu, Nelson, Li, Shern, Conger, Barette, Voss, Ding, Lu, Zhang, Beaumont, Hallacy, Koch, Gibson, Kim, Choi, McLeavey, Hesse, Fischer, Winter, Czarnecki, Jarvis, Wei, Koumouzelis, Sherburn, Kappler, Levin, Levy, Carr, Farhi, Mely, Robinson, Sasaki, Jin, Valladares, Tsipras, Li, Nguyen, Findlay, Oiwoh, Wong, Asdar, Proehl, Yang, Antonow, Kramer,
  Peterson, Sigler, Wallace, Brevdo, Mays, Khorasani, Such, Raso, Zhang, von Lohmann, Sulit, Goh, Oden, Salmon, Starace, Brockman, Salman, Bao, Hu, Wong, Wang, Schmidt, Whitney, Jun, Kirchner, de~Oliveira~Pinto, Ren, Chang, Chung, Kivlichan, O'Connell, O'Connell, Osband, Silber, Sohl, Okuyucu, Lan, Kostrikov, Sutskever, Kanitscheider, Gulrajani, Coxon, Menick, Pachocki, Aung, Betker, Crooks, Lennon, Kiros, Leike, Park, Kwon, Phang, Teplitz, Wei, Wolfe, Chen, Harris, Varavva, Lee, Shieh, Lin, Yu, Weng, Tang, Yu, Jang, Candela, Beutler, Landers, Parish, Heidecke, Schulman, Lachman, McKay, Uesato, Ward, Kim, Huizinga, Sitkin, Kraaijeveld, Gross, Kaplan, Snyder, Achiam, Jiao, Lee, Zhuang, Harriman, Fricke, Hayashi, Singhal, Shi, Karthik, Wood, Rimbach, Hsu, Nguyen, Gu-Lemberg, Button, Liu, Howe, Muthukumar, Luther, Ahmad, Kai, Itow, Workman, Pathak, Chen, Jing, Guy, Fedus, Zhou, Mamitsuka, Weng, McCallum, Held, Ouyang, Feuvrier, Zhang, Kondraciuk, Kaiser, Hewitt, Metz, Doshi, Aflak, Simens, Boyd, Thompson,
  Dukhan, Chen, Gray, Hudnall, Zhang, Aljubeh, Litwin, Zeng, Johnson, Shetty, Gupta, Shah, Yatbaz, Yang, Zhong, Glaese, Chen, Janner, Lampe, Petrov, Wu, Wang, Fradin, Pokrass, Castro, de~Castro, Pavlov, Brundage, Wang, Khan, Murati, Bavarian, Lin, Yesildal, Soto, Gimelshein, Cone, Staudacher, Summers, LaFontaine, Chowdhury, Ryder, Stathas, Turley, Tezak, Felix, Kudige, Keskar, Deutsch, Bundick, Puckett, Nachum, Okelola, Boiko, Murk, Jaffe, Watkins, Godement, Campbell-Moore, Chao, McMillan, Belov, Su, Bak, Bakkum, Deng, Dolan, Hoeschele, Welinder, Tillet, Pronin, Tillet, Dhariwal, Yuan, Dias, Lim, Arora, Troll, Lin, Lopes, Puri, Miyara, Leike, Gaubert, Zamani, Wang, Donnelly, Honsby, Smith, Sahai, Ramchandani, Huet, Carmichael, Zellers, Chen, Chen, Nigmatullin, Cheu, Jain, Altman, Schoenholz, Toizer, Miserendino, Agarwal, Culver, Ethersmith, Gray, Grove, Metzger, Hermani, Jain, Zhao, Wu, Jomoto, Wu, Shuaiqi, Xia, Phene, Papay, Narayanan, Coffey, Lee, Hall, Balaji, Broda, Stramer, Xu, Gogineni, Christianson,
  Sanders, Patwardhan, Cunninghman, Degry, Dimson, Raoux, Shadwell, Zheng, Underwood, Markov, Sherbakov, Rubin, Stasi, Kaftan, Heywood, Peterson, Walters, Eloundou, Qi, Moeller, Monaco, Kuo, Fomenko, Chang, Zheng, Zhou, Manassra, Sheu, Zaremba, Patil, Qian, Kim, Cheng, Zhang, He, Zhang, Jin, Dai, and Malkov}]{openai2024gpt4ocard}
OpenAI:~Aaron Hurst, Adam Lerer, Adam~P. Goucher, Adam Perelman, Aditya Ramesh, Aidan Clark, AJ~Ostrow, Akila Welihinda, Alan Hayes, Alec Radford, Aleksander Mądry, Alex Baker-Whitcomb, Alex Beutel, Alex Borzunov, Alex Carney, Alex Chow, Alex Kirillov, Alex Nichol, Alex Paino, and 399 others. 2024.
\newblock \href {https://arxiv.org/abs/2410.21276} {Gpt-4o system card}.
\newblock \emph{Preprint}, arXiv:2410.21276.

\bibitem[{Jaegle et~al.(2021)Jaegle, Gimeno, Brock, Vinyals, Zisserman, and Carreira}]{jaegle2021perceiver}
Andrew Jaegle, Felix Gimeno, Andy Brock, Oriol Vinyals, Andrew Zisserman, and Joao Carreira. 2021.
\newblock Perceiver: General perception with iterative attention.
\newblock In \emph{International conference on machine learning}, pages 4651--4664. PMLR.

\bibitem[{Kim et~al.(2024)Kim, Jung, Lee, and Woo}]{kim2024enclap}
Jaeyeon Kim, Jaeyoon Jung, Jinjoo Lee, and Sang~Hoon Woo. 2024.
\newblock Enclap: Combining neural audio codec and audio-text joint embedding for automated audio captioning.
\newblock In \emph{ICASSP 2024-2024 IEEE International Conference on Acoustics, Speech and Signal Processing (ICASSP)}. IEEE.

\bibitem[{Kugo et~al.(2025)Kugo, Li, Li, Gupta, Khatua, Jain, Patel, Kyuragi, Ishii, Tanabiki et~al.}]{kugo2025videomultiagents}
Noriyuki Kugo, Xiang Li, Zixin Li, Ashish Gupta, Arpandeep Khatua, Nidhish Jain, Chaitanya Patel, Yuta Kyuragi, Yasunori Ishii, Masamoto Tanabiki, and 1 others. 2025.
\newblock Videomultiagents: A multi-agent framework for video question answering.
\newblock \emph{arXiv preprint arXiv:2504.20091}.

\bibitem[{Lau et~al.(2018)Lau, Gayen, Ben~Abacha, and Demner-Fushman}]{lau2018dataset}
Jason~J Lau, Soumya Gayen, Asma Ben~Abacha, and Dina Demner-Fushman. 2018.
\newblock A dataset of clinically generated visual questions and answers about radiology images.
\newblock \emph{Scientific data}, 5(1):1--10.

\bibitem[{Lee et~al.(2024)Lee, Yoon, and Hwang}]{lee2024becotta}
Daeun Lee, Jaehong Yoon, and Sung~Ju Hwang. 2024.
\newblock Becotta: nput-dependent online blending of experts for continual test-time adaptation.
\newblock In \emph{Proceedings of the International Conference on Machine Learning (ICML)}.

\bibitem[{Lei et~al.(2024)Lei, Ge, Zhang, Sun, Yi, Shan, and Shou}]{lei2024vit}
Weixian Lei, Yixiao Ge, Jianfeng Zhang, Dylan Sun, Kun Yi, Ying Shan, and Mike~Zheng Shou. 2024.
\newblock Vit-lens: Towards omni-modal representations.
\newblock In \emph{Proceedings of the IEEE International Conference on Computer Vision and Pattern Recognition (CVPR)}.

\bibitem[{Li et~al.(2023{\natexlab{a}})Li, Zhang, Chen, Wang, Yang, and Liu}]{li2023otter}
Bo~Li, Yuanhan Zhang, Liangyu Chen, Jinghao Wang, Jingkang Yang, and Ziwei Liu. 2023{\natexlab{a}}.
\newblock Otter: A multi-modal model with in-context instruction tuning.
\newblock \emph{arXiv preprint arXiv:2305.03726}.

\bibitem[{Li et~al.(2024{\natexlab{a}})Li, Zhang, Guo, Zhang, Li, Zhang, Zhang, Zhang, Li, Liu et~al.}]{li2024llava}
Bo~Li, Yuanhan Zhang, Dong Guo, Renrui Zhang, Feng Li, Hao Zhang, Kaichen Zhang, Peiyuan Zhang, Yanwei Li, Ziwei Liu, and 1 others. 2024{\natexlab{a}}.
\newblock Llava-onevision: Easy visual task transfer.
\newblock \emph{arXiv preprint arXiv:2408.03326}.

\bibitem[{Li et~al.(2024{\natexlab{b}})Li, Tan, Qian, Li, Chaudhary, Hu, and Shen}]{li2024smoa}
Dawei Li, Zhen Tan, Peijia Qian, Yifan Li, Kumar~Satvik Chaudhary, Lijie Hu, and Jiayi Shen. 2024{\natexlab{b}}.
\newblock Smoa: Improving multi-agent large language models with sparse mixture-of-agents.
\newblock \emph{arXiv preprint arXiv:2411.03284}.

\bibitem[{Li et~al.(2024{\natexlab{c}})Li, Liu, Wu, Wang, Shen, Qu, Niu, Zhou, Huang, Li et~al.}]{li2024aria}
Dongxu Li, Yudong Liu, Haoning Wu, Yue Wang, Zhiqi Shen, Bowen Qu, Xinyao Niu, Fan Zhou, Chengen Huang, Yanpeng Li, and 1 others. 2024{\natexlab{c}}.
\newblock Aria: An open multimodal native mixture-of-experts model.
\newblock \emph{arXiv preprint arXiv:2410.05993}.

\bibitem[{Li et~al.(2023{\natexlab{b}})Li, Li, Savarese, and Hoi}]{li2023blip}
Junnan Li, Dongxu Li, Silvio Savarese, and Steven Hoi. 2023{\natexlab{b}}.
\newblock Blip-2: Bootstrapping language-image pre-training with frozen image encoders and large language models.
\newblock In \emph{International conference on machine learning}, pages 19730--19742. PMLR.

\bibitem[{Li et~al.(2022)Li, Li, Xiong, and Hoi}]{li2022blip}
Junnan Li, Dongxu Li, Caiming Xiong, and Steven Hoi. 2022.
\newblock Blip: Bootstrapping language-image pre-training for unified vision-language understanding and generation.
\newblock In \emph{International conference on machine learning}, pages 12888--12900. PMLR.

\bibitem[{Li et~al.(2025)Li, Lin, Xia, and Jin}]{li2025rethinking}
Wenzhe Li, Yong Lin, Mengzhou Xia, and Chi Jin. 2025.
\newblock Rethinking mixture-of-agents: Is mixing different large language models beneficial?
\newblock \emph{arXiv preprint arXiv:2502.00674}.

\bibitem[{Li et~al.(2020)Li, Liu, Lu, Wang, Liu, Li, and Lu}]{li2020detailed}
Yong-Lu Li, Xinpeng Liu, Han Lu, Shiyi Wang, Junqi Liu, Jiefeng Li, and Cewu Lu. 2020.
\newblock Detailed 2d-3d joint representation for human-object interaction.
\newblock In \emph{Proceedings of the IEEE International Conference on Computer Vision and Pattern Recognition (CVPR)}.

\bibitem[{Liu et~al.(2021)Liu, Zhan, Xu, Ma, Yang, and Wu}]{liu2021slake}
Bo~Liu, Li-Ming Zhan, Li~Xu, Lin Ma, Yan Yang, and Xiao-Ming Wu. 2021.
\newblock Slake: A semantically-labeled knowledge-enhanced dataset for medical visual question answering.
\newblock In \emph{2021 IEEE 18th international symposium on biomedical imaging (ISBI)}, pages 1650--1654. IEEE.

\bibitem[{Liu et~al.(2023{\natexlab{a}})Liu, Li, Wu, and Lee}]{liu2023visual}
Haotian Liu, Chunyuan Li, Qingyang Wu, and Yong~Jae Lee. 2023{\natexlab{a}}.
\newblock Visual instruction tuning.
\newblock In \emph{Advances in Neural Information Processing Systems (NeurIPS)}.

\bibitem[{Liu et~al.(2024{\natexlab{a}})Liu, Hussain, Sun, and Shan}]{liu2024music}
Shansong Liu, Atin~Sakkeer Hussain, Chenshuo Sun, and Ying Shan. 2024{\natexlab{a}}.
\newblock Music understanding llama: Advancing text-to-music generation with question answering and captioning.
\newblock In \emph{ICASSP 2024-2024 IEEE International Conference on Acoustics, Speech and Signal Processing (ICASSP)}. IEEE.

\bibitem[{Liu et~al.(2023{\natexlab{b}})Liu, Fan, Johns, Yu, Xiao, and Anandkumar}]{liu2023prismer}
Shikun Liu, Linxi Fan, Edward Johns, Zhiding Yu, Chaowei Xiao, and Anima Anandkumar. 2023{\natexlab{b}}.
\newblock Prismer: A vision-language model with an ensemble of experts.
\newblock \emph{arXiv preprint arXiv:2303.02506}.

\bibitem[{Liu et~al.(2024{\natexlab{b}})Liu, Chen, Bai, Liang, Li, Gao, and Lin}]{liu2024aligning}
Yang Liu, Weixing Chen, Yongjie Bai, Xiaodan Liang, Guanbin Li, Wen Gao, and Liang Lin. 2024{\natexlab{b}}.
\newblock Aligning cyber space with physical world: A comprehensive survey on embodied ai.
\newblock \emph{arXiv preprint arXiv:2407.06886}.

\bibitem[{Liu et~al.(2024{\natexlab{c}})Liu, Zhu, Shi, Zhang, Lou, Yang, Xi, Cao, Gu, Li et~al.}]{liu2024nvila}
Zhijian Liu, Ligeng Zhu, Baifeng Shi, Zhuoyang Zhang, Yuming Lou, Shang Yang, Haocheng Xi, Shiyi Cao, Yuxian Gu, Dacheng Li, and 1 others. 2024{\natexlab{c}}.
\newblock Nvila: Efficient frontier visual language models.
\newblock \emph{arXiv preprint arXiv:2412.04468}.

\bibitem[{Lu et~al.(2023)Lu, Peng, Cheng, Galley, Chang, Wu, Zhu, and Gao}]{lu2023chameleon}
Pan Lu, Baolin Peng, Hao Cheng, Michel Galley, Kai-Wei Chang, Ying~Nian Wu, Song-Chun Zhu, and Jianfeng Gao. 2023.
\newblock Chameleon: Plug-and-play compositional reasoning with large language models.
\newblock \emph{Advances in Neural Information Processing Systems}, 36:43447--43478.

\bibitem[{Lu et~al.(2025)Lu, Yuan, Li, Zhao, Qin, Li, Zhuo, Wen, Liu, Cao, Yan, Li, Peng, Zhang, Shi, Chen, Chen, Bai, Zhang, and Gao}]{lu2025omnicaptionercaptionerrule}
Yiting Lu, Jiakang Yuan, Zhen Li, Shitian Zhao, Qi~Qin, Xinyue Li, Le~Zhuo, Licheng Wen, Dongyang Liu, Yuewen Cao, Xiangchao Yan, Xin Li, Tianshuo Peng, Shufei Zhang, Botian Shi, Tao Chen, Zhibo Chen, Lei Bai, Bo~Zhang, and Peng Gao. 2025.
\newblock \href {https://arxiv.org/abs/2504.07089} {Omnicaptioner: One captioner to rule them all}.
\newblock \emph{Preprint}, arXiv:2504.07089.

\bibitem[{Ma et~al.(2023{\natexlab{a}})Ma, Yong, Zheng, Li, Liang, Zhu, and Huang}]{ma2022sqa3d}
Xiaojian Ma, Silong Yong, Zilong Zheng, Qing Li, Yitao Liang, Song-Chun Zhu, and Siyuan Huang. 2023{\natexlab{a}}.
\newblock Sqa3d: Situated question answering in 3d scenes.
\newblock In \emph{Proceedings of the International Conference on Learning Representations (ICLR)}.

\bibitem[{Ma et~al.(2023{\natexlab{b}})Ma, Yong, Zheng, Li, Liang, Zhu, and Huang}]{ma2023sqa3dsituatedquestionanswering}
Xiaojian Ma, Silong Yong, Zilong Zheng, Qing Li, Yitao Liang, Song-Chun Zhu, and Siyuan Huang. 2023{\natexlab{b}}.
\newblock \href {https://arxiv.org/abs/2210.07474} {Sqa3d: Situated question answering in 3d scenes}.
\newblock \emph{Preprint}, arXiv:2210.07474.

\bibitem[{Majumdar et~al.(2024)Majumdar, Ajay, Zhang, Putta, Yenamandra, Henaff, Silwal, Mcvay, Maksymets, Arnaud et~al.}]{majumdar2024openeqa}
Arjun Majumdar, Anurag Ajay, Xiaohan Zhang, Pranav Putta, Sriram Yenamandra, Mikael Henaff, Sneha Silwal, Paul Mcvay, Oleksandr Maksymets, Sergio Arnaud, and 1 others. 2024.
\newblock Openeqa: Embodied question answering in the era of foundation models.
\newblock In \emph{Proceedings of the IEEE International Conference on Computer Vision and Pattern Recognition (CVPR)}.

\bibitem[{Marino et~al.(2019)Marino, Rastegari, Farhadi, and Mottaghi}]{marino2019ok}
Kenneth Marino, Mohammad Rastegari, Ali Farhadi, and Roozbeh Mottaghi. 2019.
\newblock Ok-vqa: A visual question answering benchmark requiring external knowledge.
\newblock In \emph{Proceedings of the IEEE International Conference on Computer Vision and Pattern Recognition (CVPR)}.

\bibitem[{Mathew et~al.(2021)Mathew, Karatzas, and Jawahar}]{mathew2021docvqa}
Minesh Mathew, Dimosthenis Karatzas, and CV~Jawahar. 2021.
\newblock Docvqa: A dataset for vqa on document images.
\newblock In \emph{Proceedings of the IEEE/CVF winter conference on applications of computer vision}.

\bibitem[{Radford et~al.(2023)Radford, Kim, Xu, Brockman, McLeavey, and Sutskever}]{radford2023robust}
Alec Radford, Jong~Wook Kim, Tao Xu, Greg Brockman, Christine McLeavey, and Ilya Sutskever. 2023.
\newblock Robust speech recognition via large-scale weak supervision.
\newblock In \emph{International conference on machine learning}, pages 28492--28518. PMLR.

\bibitem[{Sakshi et~al.(2024)Sakshi, Tyagi, Kumar, Seth, Selvakumar, Nieto, Duraiswami, Ghosh, and Manocha}]{sakshi2024mmaumassivemultitaskaudio}
S~Sakshi, Utkarsh Tyagi, Sonal Kumar, Ashish Seth, Ramaneswaran Selvakumar, Oriol Nieto, Ramani Duraiswami, Sreyan Ghosh, and Dinesh Manocha. 2024.
\newblock \href {https://arxiv.org/abs/2410.19168} {Mmau: A massive multi-task audio understanding and reasoning benchmark}.
\newblock \emph{Preprint}, arXiv:2410.19168.

\bibitem[{Shazeer et~al.(2017)Shazeer, Mirhoseini, Maziarz, Davis, Le, Hinton, and Dean}]{shazeer2017outrageously}
Noam Shazeer, Azalia Mirhoseini, Krzysztof Maziarz, Andy Davis, Quoc Le, Geoffrey Hinton, and Jeff Dean. 2017.
\newblock Outrageously large neural networks: The sparsely-gated mixture-of-experts layer.
\newblock \emph{arXiv preprint arXiv:1701.06538}.

\bibitem[{Sur{\'\i}s et~al.(2023)Sur{\'\i}s, Menon, and Vondrick}]{suris2023vipergpt}
D{\'\i}dac Sur{\'\i}s, Sachit Menon, and Carl Vondrick. 2023.
\newblock Vipergpt: Visual inference via python execution for reasoning.
\newblock In \emph{Proceedings of the IEEE/CVF international conference on computer vision}, pages 11888--11898.

\bibitem[{Tan and Bansal(2019)}]{tan2019lxmert}
Hao Tan and Mohit Bansal. 2019.
\newblock Lxmert: Learning cross-modality encoder representations from transformers.
\newblock In \emph{Proceedings of the 2019 Conference on Empirical Methods in Natural Language Processing and the 9th International Joint Conference on Natural Language Processing (EMNLP-IJCNLP)}, pages 5100--5111.

\bibitem[{Tanaka et~al.(2023)Tanaka, Nishida, Nishida, Hasegawa, Saito, and Saito}]{tanaka2023slidevqa}
Ryota Tanaka, Kyosuke Nishida, Kosuke Nishida, Taku Hasegawa, Itsumi Saito, and Kuniko Saito. 2023.
\newblock Slidevqa: A dataset for document visual question answering on multiple images.
\newblock In \emph{Proceedings of the AAAI National Conference on Artificial Intelligence (AAAI)}.

\bibitem[{Tang et~al.(2024)Tang, Yu, Sun, Chen, Tan, Li, Lu, Ma, and Zhang}]{tang2023salmonn}
Changli Tang, Wenyi Yu, Guangzhi Sun, Xianzhao Chen, Tian Tan, Wei Li, Lu~Lu, Zejun Ma, and Chao Zhang. 2024.
\newblock Salmonn: Towards generic hearing abilities for large language models.
\newblock In \emph{Proceedings of the International Conference on Learning Representations (ICLR)}.

\bibitem[{Team(2024)}]{team2024chameleon}
Chameleon Team. 2024.
\newblock Chameleon: Mixed-modal early-fusion foundation models.
\newblock \emph{arXiv preprint arXiv:2405.09818}.

\bibitem[{Team et~al.(2024)Team, Georgiev, Lei, Burnell, Bai, Gulati, Tanzer, Vincent, Pan, Wang, Mariooryad, Ding, Geng, Alcober, Frostig, Omernick, Walker, Paduraru, Sorokin, Tacchetti, Gaffney, Daruki, Sercinoglu, Gleicher, Love, Voigtlaender, Jain, Surita, Mohamed, Blevins, Ahn, Zhu, Kawintiranon, Firat, Gu, Zhang, Rahtz, Faruqui, Clay, Gilmer, Co-Reyes, Penchev, Zhu, Morioka, Hui, Haridasan, Campos, Mahdieh, Guo, Hassan, Kilgour, Vezer, Cheng, de~Liedekerke, Goyal, Barham, Strouse, Noury, Adler, Sundararajan, Vikram, Lepikhin, Paganini, Garcia, Yang, Valter, Trebacz, Vodrahalli, Asawaroengchai, Ring, Kalb, Soares, Brahma, Steiner, Yu, Mentzer, He, Gonzalez, Xu, Kaufman, Shafey, Oh, Hennigan, van~den Driessche, Odoom, Lucic, Roelofs, Lall, Marathe, Chan, Ontanon, He, Teplyashin, Lai, Crone, Damoc, Ho, Riedel, Lenc, Yeh, Chowdhery, Xu, Kazemi, Amid, Petrushkina, Swersky, Khodaei, Chen, Larkin, Pinto, Yan, Badia, Patil, Hansen, Orr, Arnold, Grimstad, Dai, Douglas, Sinha, Yadav, Chen, Gribovskaya, Austin,
  Zhao, Patel, Komarek, Austin, Borgeaud, Friso, Goyal, Caine, Cao, Chung, Lamm, Barth-Maron, Kagohara, Olszewska, Chen, Shivakumar, Agarwal, Godhia, Rajwar, Snaider, Dotiwalla, Liu, Barua, Ungureanu, Zhang, Batsaikhan, Wirth, Qin, Danihelka, Doshi, Chadwick, Chen, Jain, Le, Kar, Gurumurthy, Li, Sang, Liu, Lamprou, Munoz, Lintz, Mehta, Howard, Reynolds, Aroyo, Wang, Blanco, Cassirer, Griffith, Das, Lee, Sygnowski, Fisher, Besley, Powell, Ahmed, Paulus, Reitter, Borsos, Joshi, Pope, Hand, Selo, Jain, Sethi, Goel, Makino, May, Yang, Schalkwyk, Butterfield, Hauth, Goldin, Hawkins, Senter, Brin, Woodman, Ritter, Noland, Giang, Bolina, Lee, Blyth, Mackinnon, Reid, Sarvana, Silver, Chen, Wang, Maggiore, Chang, Attaluri, Thornton, Chiu, Bunyan, Levine, Chung, Eltyshev, Si, Lillicrap, Brady, Aggarwal, Wu, Xu, McIlroy, Badola, Sandhu, Moreira, Stokowiec, Hemsley, Li, Tudor, Shyam, Rahimtoroghi, Haykal, Sprechmann, Zhou, Mincu, Li, Addanki, Krishna, Wu, Frechette, Eyal, Dafoe, Lacey, Whang, Avrahami, Zhang, Taropa,
  Lin, Toyama, Rutherford, Sano, Choe, Tomala, Safranek-Shrader, Kassner, Pajarskas, Harvey, Sechrist, Fortunato, Lyu, Elsayed, Kuang, Lottes, Chu, Jia, Chen, Humphreys, Baumli, Tao, Samuel, dos Santos, Andreassen, Rakićević, Grewe, Kumar, Winkler, Caton, Brock, Dalmia, Sheahan, Barr, Miao, Natsev, Devlin, Behbahani, Prost, Sun, Myaskovsky, Pillai, Hurt, Lazaridou, Xiong, Zheng, Pardo, Li, Horgan, Stanton, Ambar, Xia, Lince, Wang, Mustafa, Webson, Lee, Anil, Wicke, Dozat, Sinha, Piqueras, Dabir, Upadhyay, Boral, Hendricks, Fry, Djolonga, Su, Walker, Labanowski, Huang, Misra, Chen, Skerry-Ryan, Singh, Rijhwani, Yu, Castro-Ros, Changpinyo, Datta, Bagri, Hrafnkelsson, Maggioni, Zheng, Sulsky, Hou, Paine, Yang, Riesa, Rogozinska, Marcus, Badawy, Zhang, Wang, Miller, Greer, Sjos, Nova, Zen, Chaabouni, Rosca, Jiang, Chen, Liu, Sainath, Krikun, Polozov, Lespiau, Newlan, Cankara, Kwak, Xu, Chen, Coenen, Meyer, Tsihlas, Ma, Gottweis, Xing, Gu, Miao, Frank, Cankara, Ganapathy, Dasgupta, Hughes-Fitt, Chen, Reid, Rong,
  Fan, van Amersfoort, Zhuang, Cohen, Gu, Mohananey, Ilic, Tobin, Wieting, Bortsova, Thacker, Wang, Caveness, Chiu, Sezener, Kaskasoli, Baker, Millican, Elhawaty, Aisopos, Lebsack, Byrd, Dai, Jia, Wiethoff, Davoodi, Weston, Yagati, Ahuja, Gao, Pundak, Zhang, Azzam, Sim, Caelles, Keeling, Sharma, Swing, Li, Liu, Bostock, Bansal, Nado, Anand, Lipschultz, Karmarkar, Proleev, Ittycheriah, Yeganeh, Polovets, Faust, Sun, Rrustemi, Li, Shivanna, Liu, Welty, Lebron, Baddepudi, Krause, Parisotto, Soricut, Xu, Bloxwich, Johnson, Neyshabur, Mao-Jones, Wang, Ramasesh, Abbas, Guez, Segal, Nguyen, Svensson, Hou, York, Milan, Bridgers, Gworek, Tagliasacchi, Lee-Thorp, Chang, Guseynov, Hartman, Kwong, Zhao, Kashem, Cole, Miech, Tanburn, Phuong, Pavetic, Cevey, Comanescu, Ives, Yang, Du, Li, Zhang, Iinuma, Hu, Roy, Bijwadia, Zhu, Martins, Saputro, Gergely, Zheng, Jia, Antonoglou, Sadovsky, Gu, Bi, Andreev, Samangooei, Khan, Kocisky, Filos, Kumar, Bishop, Yu, Hodkinson, Mittal, Shah, Moufarek, Cheng, Bloniarz, Lee, Pejman,
  Michel, Spencer, Feinberg, Xiong, Savinov, Smith, Shakeri, Tran, Chesus, Bohnet, Tucker, von Glehn, Muir, Mao, Kazawa, Slone, Soparkar, Shrivastava, Cobon-Kerr, Sharman, Pavagadhi, Araya, Misiunas, Ghelani, Laskin, Barker, Li, Briukhov, Houlsby, Glaese, Lakshminarayanan, Schucher, Tang, Collins, Lim, Feng, Recasens, Lai, Magni, Cao, Siddhant, Ashwood, Orbay, Dehghani, Brennan, He, Xu, Gao, Saroufim, Molloy, Wu, Arnold, Chang, Schrittwieser, Buchatskaya, Radpour, Polacek, Giordano, Bapna, Tokumine, Hellendoorn, Sottiaux, Cogan, Severyn, Saleh, Thakoor, Shefey, Qiao, Gaba, yiin Chang, Swanson, Zhang, Lee, Rubenstein, Song, Kwiatkowski, Koop, Kannan, Kao, Schuh, Stjerngren, Ghiasi, Gibson, Vilnis, Yuan, Ferreira, Kamath, Klimenko, Franko, Xiao, Bhattacharya, Patel, Wang, Morris, Strudel, Sharma, Choy, Hashemi, Landon, Finkelstein, Jhakra, Frye, Barnes, Mauger, Daun, Baatarsukh, Tung, Farhan, Michalewski, Viola, de~Chaumont~Quitry, Lan, Hudson, Wang, Fischer, Zheng, White, Dragan, baptiste Alayrac, Ni, Pritzel,
  Iwanicki, Isard, Bulanova, Zilka, Dyer, Sachan, Srinivasan, Muckenhirn, Cai, Mandhane, Tariq, Rae, Wang, Ayoub, FitzGerald, Zhao, Han, Alberti, Garrette, Krishnakumar, Gimenez, Levskaya, Sohn, Matak, Iturrate, Chang, Xiang, Cao, Ranka, Brown, Hutter, Mirrokni, Chen, Yao, Egyed, Galilee, Liechty, Kallakuri, Palmer, Ghemawat, Liu, Tao, Thornton, Green, Jasarevic, Lin, Cotruta, Tan, Fiedel, Yu, Chi, Neitz, Heitkaemper, Sinha, Zhou, Sun, Kaed, Hulse, Mishra, Georgaki, Kudugunta, Farabet, Shafran, Vlasic, Tsitsulin, Ananthanarayanan, Carin, Su, Sun, V, Carvajal, Broder, Comsa, Repina, Wong, Chen, Hawkins, Filonov, Loher, Hirnschall, Wang, Ye, Burns, Cate, Wright, Piccinini, Zhang, Lin, Gog, Kulizhskaya, Sreevatsa, Song, Cobo, Iyer, Tekur, Garrido, Xiao, Kemp, Zheng, Li, Agarwal, Ngani, Goshvadi, Santamaria-Fernandez, Fica, Chen, Gorgolewski, Sun, Garg, Ye, Eslami, Hua, Simon, Joshi, Kim, Tenney, Potluri, Thiet, Yuan, Luisier, Chronopoulou, Scellato, Srinivasan, Chen, Koverkathu, Dalibard, Xu, Saeta, Anderson,
  Sellam, Fernando, Huot, Jung, Varadarajan, Quinn, Raul, Le, Habalov, Clark, Jalan, Bullard, Singhal, Luong, Wang, Rajayogam, Eisenschlos, Jia, Finchelstein, Yakubovich, Balle, Fink, Agarwal, Li, Dvijotham, Pal, Kang, Konzelmann, Beattie, Dousse, Wu, Crocker, Elkind, Jonnalagadda, Lee, Holtmann-Rice, Kallarackal, Liu, Vnukov, Vats, Invernizzi, Jafari, Zhou, Taylor, Prendki, Wu, Eccles, Liu, Kopparapu, Beaufays, Angermueller, Marzoca, Sarcar, Dib, Stanway, Perbet, Trdin, Sterneck, Khorlin, Li, Wu, Goenka, Madras, Goldshtein, Gierke, Zhou, Liu, Liang, White, Li, Singh, Bahargam, Epstein, Basu, Lao, Ozturel, Crous, Zhai, Lu, Tung, Gaur, Walton, Dixon, Zhang, Globerson, Uy, Bolt, Wiles, Nasr, Shumailov, Selvi, Piccinno, Aguilar, McCarthy, Khalman, Shukla, Galic, Carpenter, Villela, Zhang, Richardson, Martens, Bosnjak, Belle, Seibert, Alnahlawi, McWilliams, Singh, Louis, Ding, Popovici, Simicich, Knight, Mehta, Gupta, Shi, Fatehi, Mitrovic, Grills, Pagadora, Munkhdalai, Petrova, Eisenbud, Zhang, Yates, Mittal,
  Tripuraneni, Assael, Brovelli, Jain, Velimirovic, Akbulut, Mu, Macherey, Kumar, Xu, Qureshi, Comanici, Wiesner, Gong, Ruddock, Bauer, Felt, GP, Arnab, Zelle, Rothfuss, Rosgen, Shenoy, Seybold, Li, Mudigonda, Erdogan, Xia, Simsa, Michi, Yao, Yew, Kan, Caswell, Radebaugh, Elisseeff, Valenzuela, McKinney, Paterson, Cui, Latorre-Chimoto, Kim, Zeng, Durden, Ponnapalli, Sosea, Choquette-Choo, Manyika, Robenek, Vashisht, Pereira, Lam, Velic, Owusu-Afriyie, Lee, Bolukbasi, Parrish, Lu, Park, Venkatraman, Talbert, Rosique, Cheng, Sozanschi, Paszke, Kumar, Austin, Li, Salama, Perz, Kim, Dukkipati, Baryshnikov, Kaplanis, Sheng, Chervonyi, Unlu, de~Las~Casas, Askham, Tunyasuvunakool, Gimeno, Poder, Kwak, Miecnikowski, Mirrokni, Dimitriev, Parisi, Liu, Tsai, Shevlane, Kouridi, Garmon, Goedeckemeyer, Brown, Vijayakumar, Elqursh, Jazayeri, Huang, Carthy, Hoover, Kim, Kumar, Chen, Biles, Bingham, Rosen, Wang, Tan, Engel, Pongetti, de~Cesare, Hwang, Yu, Pullman, Narayanan, Levin, Gopal, Li, Aharoni, Trinh, Lo, Casagrande,
  Vij, Matthey, Ramadhana, Matthews, Carey, Johnson, Goranova, Shah, Ashraf, Dasgupta, Larsen, Wang, Vuyyuru, Jiang, Ijazi, Osawa, Smith, Boppana, Bilal, Koizumi, Xu, Altun, Shabat, Bariach, Korchemniy, Choo, Ronneberger, Iwuanyanwu, Zhao, Soergel, Hsieh, Cai, Iqbal, Sundermeyer, Chen, Bursztein, Malaviya, Biadsy, Shroff, Dhillon, Latkar, Dyer, Forbes, Nicosia, Nikolaev, Greene, Georgiev, Wang, Martin, Sedghi, Zhang, Banzal, Fritz, Rao, Wang, Zhang, Patraucean, Du, Mordatch, Jurin, Liu, Dubey, Mohan, Nowakowski, Ion, Wei, Tojo, Raad, Hudson, Keshava, Agrawal, Ramirez, Wu, Nguyen, Liu, Sewak, Petrini, Choi, Philips, Wang, Bica, Garg, Wilkiewicz, Agrawal, Li, Guo, Xue, Shaik, Leach, Khan, Wiesinger, Jerome, Chakladar, Wang, Ornduff, Abu, Ghaffarkhah, Wainwright, Cortes, Liu, Maynez, Terzis, Samangouei, Mansour, Kępa, Aubet, Algymr, Banica, Weisz, Orban, Senges, Andrejczuk, Geller, Santo, Anklin, Merey, Baeuml, Strohman, Bai, Petrov, Wu, Hassabis, Kavukcuoglu, Dean, and
  Vinyals}]{geminiteam2024gemini15unlockingmultimodal}
Gemini Team, Petko Georgiev, Ving~Ian Lei, Ryan Burnell, Libin Bai, Anmol Gulati, Garrett Tanzer, Damien Vincent, Zhufeng Pan, Shibo Wang, Soroosh Mariooryad, Yifan Ding, Xinyang Geng, Fred Alcober, Roy Frostig, Mark Omernick, Lexi Walker, Cosmin Paduraru, Christina Sorokin, and 1118 others. 2024.
\newblock \href {https://arxiv.org/abs/2403.05530} {Gemini 1.5: Unlocking multimodal understanding across millions of tokens of context}.
\newblock \emph{Preprint}, arXiv:2403.05530.

\bibitem[{Wang et~al.(2024)Wang, Wang, Athiwaratkun, Zhang, and Zou}]{wang2024mixture}
Junlin Wang, Jue Wang, Ben Athiwaratkun, Ce~Zhang, and James Zou. 2024.
\newblock Mixture-of-agents enhances large language model capabilities.
\newblock \emph{arXiv preprint arXiv:2406.04692}.

\bibitem[{Wang et~al.(2025)Wang, Yu, Stengel-Eskin, Yoon, Cheng, Bertasius, and Bansal}]{wang2025videotree}
Ziyang Wang, Shoubin Yu, Elias Stengel-Eskin, Jaehong Yoon, Feng Cheng, Gedas Bertasius, and Mohit Bansal. 2025.
\newblock Videotree: Adaptive tree-based video representation for llm reasoning on long videos.
\newblock In \emph{Proceedings of the IEEE Conference on Computer Vision and Pattern Recognition}.

\bibitem[{Winata et~al.(2024)Winata, Hudi, Irawan, Anugraha, Putri, Wang, Nohejl, Prathama, Ousidhoum, Amriani et~al.}]{winata2024worldcuisines}
Genta~Indra Winata, Frederikus Hudi, Patrick~Amadeus Irawan, David Anugraha, Rifki~Afina Putri, Yutong Wang, Adam Nohejl, Ubaidillah~Ariq Prathama, Nedjma Ousidhoum, Afifa Amriani, and 1 others. 2024.
\newblock Worldcuisines: A massive-scale benchmark for multilingual and multicultural visual question answering on global cuisines.
\newblock \emph{arXiv preprint arXiv:2410.12705}.

\bibitem[{Xu et~al.(2025)Xu, Guo, He, Hu, He, Bai, Chen, Wang, Fan, Dang et~al.}]{xu2025qwen2}
Jin Xu, Zhifang Guo, Jinzheng He, Hangrui Hu, Ting He, Shuai Bai, Keqin Chen, Jialin Wang, Yang Fan, Kai Dang, and 1 others. 2025.
\newblock Qwen2. 5-omni technical report.
\newblock \emph{arXiv preprint arXiv:2503.20215}.

\bibitem[{Yao et~al.(2024)Yao, Yu, Zhang, Wang, Cui, Zhu, Cai, Li, Zhao, He et~al.}]{yao2024minicpm}
Yuan Yao, Tianyu Yu, Ao~Zhang, Chongyi Wang, Junbo Cui, Hongji Zhu, Tianchi Cai, Haoyu Li, Weilin Zhao, Zhihui He, and 1 others. 2024.
\newblock Minicpm-v: A gpt-4v level mllm on your phone.
\newblock \emph{arXiv preprint arXiv:2408.01800}.

\bibitem[{Yu et~al.(2024)Yu, Yoon, and Bansal}]{yu2024crema}
Shoubin Yu, Jaehong Yoon, and Mohit Bansal. 2024.
\newblock Crema: Generalizable and efficient video-language reasoning via multimodal modular fusion.
\newblock \emph{arXiv preprint arXiv:2402.05889}.

\bibitem[{Zellers et~al.(2022)Zellers, Lu, Lu, Yu, Zhao, Salehi, Kusupati, Hessel, Farhadi, and Choi}]{zellers2022merlot}
Rowan Zellers, Jiasen Lu, Ximing Lu, Youngjae Yu, Yanpeng Zhao, Mohammadreza Salehi, Aditya Kusupati, Jack Hessel, Ali Farhadi, and Yejin Choi. 2022.
\newblock Merlot reserve: Neural script knowledge through vision and language and sound.
\newblock In \emph{Proceedings of the IEEE International Conference on Computer Vision and Pattern Recognition (CVPR)}.

\bibitem[{Zhang et~al.(2025)Zhang, Lin, Wang, Bansal, and Bertasius}]{zhang2025silvr}
Ce~Zhang, Yan-Bo Lin, Ziyang Wang, Mohit Bansal, and Gedas Bertasius. 2025.
\newblock Silvr: A simple language-based video reasoning framework.
\newblock \emph{arXiv preprint arXiv:2505.24869}.

\bibitem[{Zhang et~al.(2023)Zhang, Cai, Chen, Zhang, Zhang, Chen, Chang, Wang, and Liu}]{zhang2023robust}
Yihua Zhang, Ruisi Cai, Tianlong Chen, Guanhua Zhang, Huan Zhang, Pin-Yu Chen, Shiyu Chang, Zhangyang Wang, and Sijia Liu. 2023.
\newblock Robust mixture-of-expert training for convolutional neural networks.
\newblock In \emph{Proceedings of the IEEE/CVF International Conference on Computer Vision}, pages 90--101.

\bibitem[{Zhang et~al.(2024{\natexlab{a}})Zhang, Wu, Li, Li, Ma, Liu, and Li}]{zhang2024video}
Yuanhan Zhang, Jinming Wu, Wei Li, Bo~Li, Zejun Ma, Ziwei Liu, and Chunyuan Li. 2024{\natexlab{a}}.
\newblock Video instruction tuning with synthetic data.
\newblock \emph{arXiv preprint arXiv:2410.02713}.

\bibitem[{Zhang et~al.(2024{\natexlab{b}})Zhang, Colman, Guo, Shahriyari, and Bharaj}]{zhang2024common}
Yue Zhang, Ben Colman, Xiao Guo, Ali Shahriyari, and Gaurav Bharaj. 2024{\natexlab{b}}.
\newblock Common sense reasoning for deepfake detection.
\newblock In \emph{European Conference on Computer Vision}, pages 399--415. Springer.

\bibitem[{Zhang et~al.(2024{\natexlab{c}})Zhang, Ma, Li, Qiao, Wang, Chai, Wu, Bansal, and Kordjamshidi}]{zhang2024vision}
Yue Zhang, Ziqiao Ma, Jialu Li, Yanyuan Qiao, Zun Wang, Joyce Chai, Qi~Wu, Mohit Bansal, and Parisa Kordjamshidi. 2024{\natexlab{c}}.
\newblock Vision-and-language navigation today and tomorrow: A survey in the era of foundation models.
\newblock \emph{arXiv preprint arXiv:2407.07035}.

\bibitem[{Zhang et~al.(2024{\natexlab{d}})Zhang, Xu, Shen, Kordjamshidi, and Huang}]{zhang2024spartun3d}
Yue Zhang, Zhiyang Xu, Ying Shen, Parisa Kordjamshidi, and Lifu Huang. 2024{\natexlab{d}}.
\newblock Spartun3d: Situated spatial understanding of 3d world in large language models.
\newblock \emph{arXiv preprint arXiv:2410.03878}.

\bibitem[{Zhou et~al.(2022)Zhou, Lei, Liu, Du, Huang, Zhao, Dai, Le, Laudon et~al.}]{zhou2022mixture}
Yanqi Zhou, Tao Lei, Hanxiao Liu, Nan Du, Yanping Huang, Vincent Zhao, Andrew~M Dai, Quoc~V Le, James Laudon, and 1 others. 2022.
\newblock Mixture-of-experts with expert choice routing.
\newblock \emph{Advances in Neural Information Processing Systems}, 35:7103--7114.

\bibitem[{Zhou et~al.(2025)Zhou, He, Su, Han, Jang, Bertasius, Bansal, and Yao}]{zhou2025reagent}
Yiyang Zhou, Yangfan He, Yaofeng Su, Siwei Han, Joel Jang, Gedas Bertasius, Mohit Bansal, and Huaxiu Yao. 2025.
\newblock Reagent-v: A reward-driven multi-agent framework for video understanding.
\newblock \emph{arXiv preprint arXiv:2506.01300}.

\bibitem[{Zoph et~al.(2022)Zoph, Bello, Kumar, Du, Huang, Dean, Shazeer, and Fedus}]{zoph2022st}
Barret Zoph, Irwan Bello, Sameer Kumar, Nan Du, Yanping Huang, Jeff Dean, Noam Shazeer, and William Fedus. 2022.
\newblock St-moe: Designing stable and transferable sparse expert models.
\newblock \emph{arXiv preprint arXiv:2202.08906}.

\end{thebibliography}
}

\clearpage

\appendix
\section{Appendix}
\label{sec:appendix}

\subsection{Prompts for Router and Aggregator}
Fig.~\ref{fig:expert-prompt} and Fig.~\ref{fig:aggreagator} illustrate the prompt designs used in our framework, respectively. Fig.~\ref{fig:expert-prompt} presents the expert selection prompt, which identifies skill-specialized experts for the input task context and question. Fig.~\ref{fig:aggreagator} shows the aggregator prompt, which integrates the outputs from the selected experts and guides LRM to reason over them and generate the final answer. 

To generate captions from different experts that explicitly focus on specialized skills, we design and apply skill-specific prompts for each expert. For instance, to obtain detailed image descriptions, we prompt Omnicaptioner-QWen2.5-7B with \textit{``You are a helpful assistant focused on providing detailed descriptions and background information for images. Analyze the given image and generate a comprehensive caption that includes the visual style, spatial relationships between elements, texture details, descriptions of the main objects, and relevant world knowledge to enhance understanding."} whereas, for concise image summaries, we use the prompt \textit{``You are a helpful assistant focused on creating short captions for images. Analyze the provided image and generate a concise caption that highlights the main subject."} Similarly, for other experts, we provide clear and direct prompts explicitly highlighting the required skill.

\subsection{Evaluation Datasets}
We validate our framework across various challenging multimodal tasks, including Video Reasoning (Video-MMMU \citep{hu2025videommmuevaluatingknowledgeacquisition}), Audio QA (MMAU \citep{sakshi2024mmaumassivemultitaskaudio}), 3D Situated Reasoning (SQA3D \citep{ma2023sqa3dsituatedquestionanswering}), and Medical QA (M3D \citep{bai2024m3d}). We specifically choose these benchmarks because they represent diverse reasoning complexities, modality interactions, and practical application scenarios.

\noindent\textbf{Video-MMMU} \citep{hu2025videommmuevaluatingknowledgeacquisition} provides a robust evaluation of models' abilities to integrate multimodal educational content and reason across diverse knowledge domains from educational videos. 
In our evaluation, we test on the full $900$ video reasoning questions. 

\noindent\textbf{MMAU} \citep{sakshi2024mmaumassivemultitaskaudio}
 is a benchmark designed to evaluate multimodal audio understanding with curated audio clips paired with human-annotated natural language questions spanning speech, environmental sounds, and music. Evaluation is conducted on a $1$K validation set with available ground truth annotations.
 
\noindent\textbf{SQA3D} \citep{ma2023sqa3dsituatedquestionanswering} is a benchmark specifically designed for evaluating situated 3D scene understanding. It presents situations such as \textit{"You are standing beside a table"}, requiring models to reason about 3D spatial relationships between the described viewpoint and surrounding objects. In our evaluation, we use the SQA3D test set, which includes around $3$K human-annotated question-answer pairs.

\noindent\textbf{M3D-VQA}~\citep{bai2024m3d} is a benchmark designed for expert-level reasoning over medical data, specifically focusing on 3D CT scans. It consists of natural-language question-answer pairs covering five diagnostic dimensions: plane, phase, organ, abnormality, and location. For evaluation, we sample $500$ question-answer pairs under the closed-ended QA setting.

\subsection{Running Efficiency Evaluation}

In~\cref{tab:efficiency}, we provide detailed computational cost analysis, including: average expert skills calls and running time (router, captioner and aggregator) as listed in Table A. We observe that Video-MMMU requires, on average, more skills than other tasks. This is because video tasks contain richer multimodal information and therefore need more experts or skills to handle effectively. Additionally, our system allows a flexible trade-off between running time and accuracy by adjusting the size of the expert pool in the router prompt.

\begin{table}[]
    \small
    \centering
    \begin{tabular}{lcc}
        \toprule
       \textbf{Dataset}  & \textbf{Avg. Skills Call} & \textbf{Avg. Running Time} \\ \midrule
    VideoMMMU &	2.10	&21.10 \\
SQA3D	&1.81&	18.03 \\
MMAU&	1.10&	13.31 \\
M3D&	1.10	&13.81  \\ \bottomrule
    \end{tabular}
    \caption{Average skill calls (times) and running time (s) across datasets/benchmarks.}
    \label{tab:efficiency}
\end{table}

\subsection{Comparison With Other Agent Systems}

In~\cref{tab:other}, we provide extra results with the recent multimodal multi-agent system. Our MEXA achieves better performance on Video-MMMU and shows more generalizability over different tasks, while other methods are limited to the specific task.

\begin{table*}[]
    \small
    \centering
    \begin{tabular}{lcc}
        \toprule
\textbf{Method}  & \textbf{Video-MMMU} & \textbf{MMAU} \\ \midrule
VideoMultiAgent~\cite{kugo2025videomultiagents} &	$62.88$	&not applicable
 \\
ReAgent-V~\cite{zhou2025reagent}	& $61.11$ &	not applicable
 \\
MEXA &	$71.47$ &	$45.9$ \\
 \bottomrule
    \end{tabular}
    \caption{Comparison with other Agent-based works.}
    \label{tab:other}
\end{table*}

\subsection{Extra Discussion on Expert Overlapping and Conflict}

We clarify that our framework addresses redundancy and conflicts in both expert selection and aggregation. Specifically, our MLLM router dynamically selects experts based on their semantic relevance to the given query, significantly reducing unnecessary redundancy from activating all available experts.
Additionally, our aggregator leverages contextual reasoning to integrate expert outputs, effectively resolving conflicts by prioritizing information consistency and relevance.  

We use two examples to show how our model addresses redundancy and conflicts, respectively.

\textbf{Redundancy}: Consider the query: "Describe the spatial relationship between the table and the chair." Suppose both the 3D Expert and Visual Expert independently provide the response: "The chair is placed next to the table on its left side." In this scenario, our aggregator effectively integrates these overlapping outputs, boosting confidence in the identified spatial relationship ("left") between the table and chair.

\textbf{Conflicts}: Consider the query: "Which object is closer to the door, the chair or the table?" In this scenario, suppose two different experts provide conflicting outputs. The 3D scan expert generates: "The chair is closer to the door.", while the visual expert generates: "The table appears closer to the door." This exemplifies a conflict where two experts offer totally opposing answers. Our aggregator resolves such conflicts by analyzing the query and prioritizes the 3D scan expert, as spatial depth is more accurately captured through 3D geometry than 2D visual appearance.

\subsection{Qualitative Examples}
\begin{figure*}[t]
  \centering
  \includegraphics[width=0.9\linewidth]{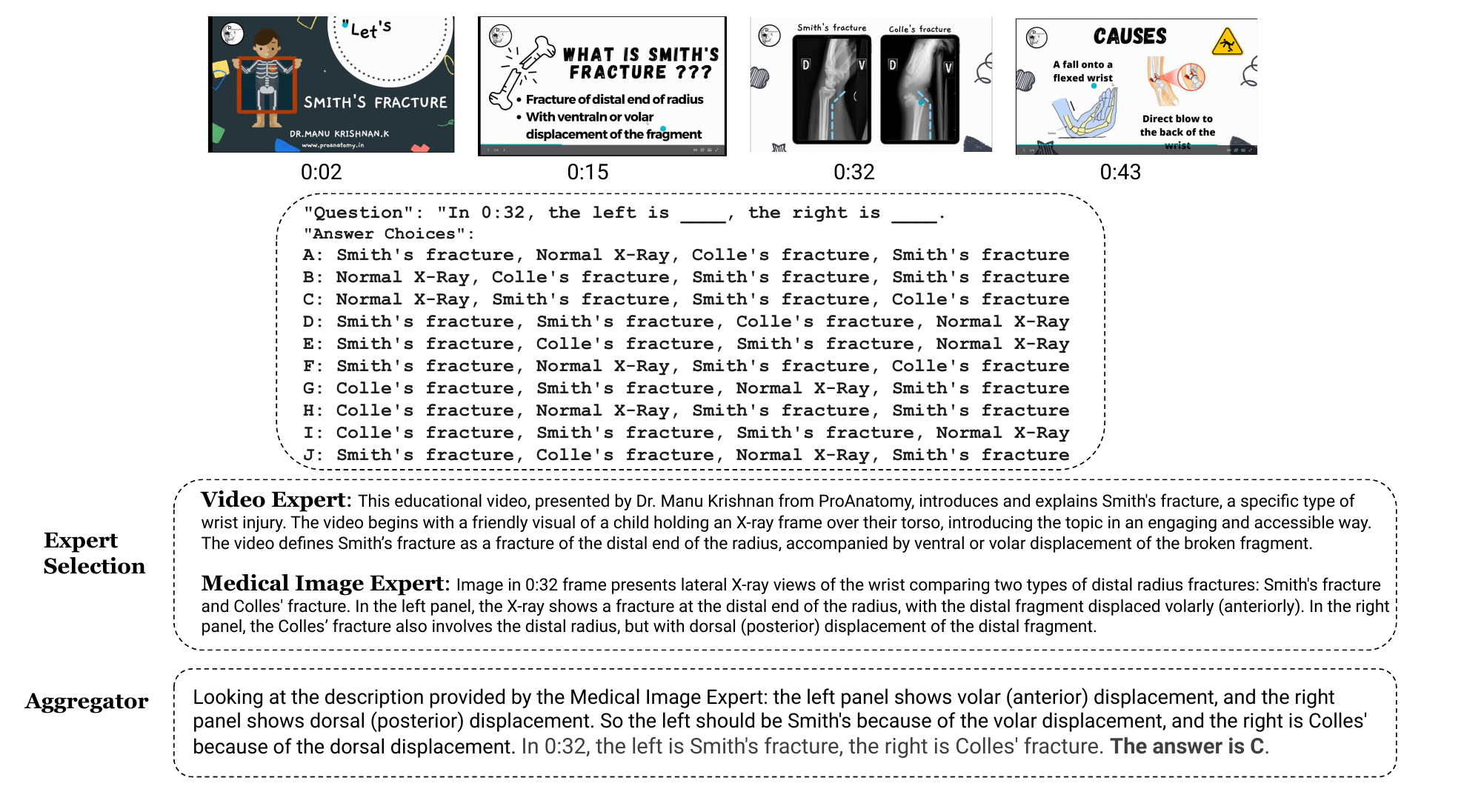}
  \caption{A qualitative example of Video-MMMU.}
  \label{fig:vidommmu}

  \vspace{1em}  

  \includegraphics[width=0.8\linewidth]{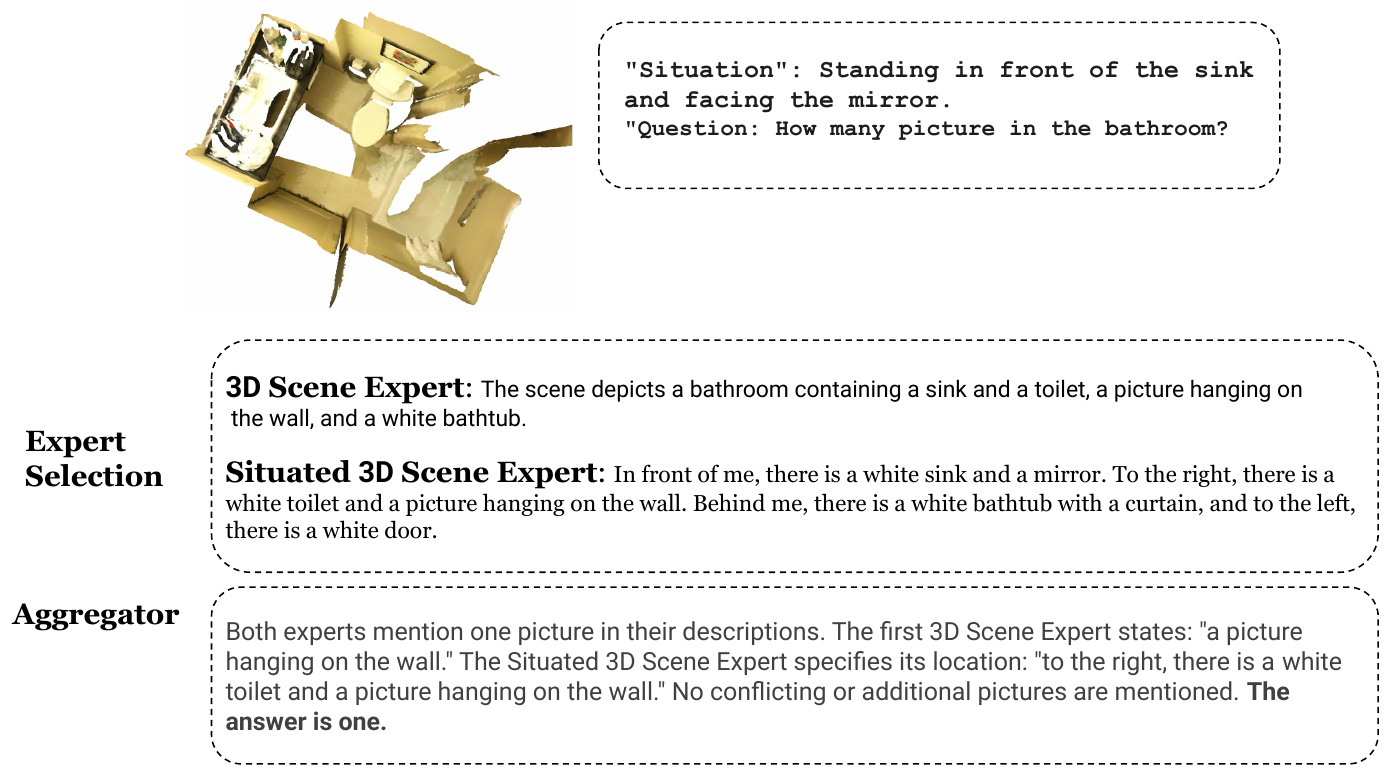}
  \caption{A qualitative example of SQA3D.}
  \label{fig:sqa3d}

\vspace{1em}

\begin{tcolorbox}[title=Prompts for Aggregator, colback=gray!5, colframe=gray!80, fonttitle=\bfseries]
\small
You are an answerer for a video question answering, audio question answering, 3D situated question answering, or medical visual question answering.
Below is information provided by multiple expert modules relevant to solving the question:

\begin{itemize}
    \item \textbf{Expert 1:}  \{Text Descriptions from Expert 1\}
    \item \textbf{Expert 2:}  \{Text Descriptions from Expert 2\}
    \item \textbf{Expert 3:}  \{Text Descriptions from Expert 3\}
    \item  ...
\end{itemize}

Using the information above, please select the best answer to the question and provide a brief explanation if needed.

\textbf{Question:} \texttt{"\{\}"}

\end{tcolorbox}
\caption{Prompts for Aggregator.}
\label{fig:aggreagator}
\end{figure*}

Fig~\ref{fig:vidommmu}  and Fig~\ref{fig:sqa3d} show two qualitative examples from Video-MMMU and SQA3D, respectively.
In the Video-MMMU example, our framework effectively selects the most relevant experts, including the video expert and the medical image expert. The aggregator then filters and prioritizes key information extracted by the medical image expert, allowing the reasoning module to accurately fill in the missing information required to answer the question. In the SQA3D example, both the general 3D scene expert and the situated 3D scene expert are activated. Their outputs are jointly considered by the aggregator to produce a coherent answer.

\begin{figure*}[t]
\centering
\small
\begin{tcolorbox}[title=Prompts for Expert Selection, colback=gray!5, colframe=gray!80, fonttitle=\bfseries]

You are an expert multimodal reasoning assistant. Given a multimodal question (e.g., related to video, audio, 3D scenes, medical images, etc.), your task is to select all relevant skills and modalities required to accurately answer the question.

\textbf{Task Type:} \texttt{"\{\}"} \\
\textbf{Question:} \texttt{"\{\}"} \\
\textbf{Options:} \texttt{\{\}} \\
Available Skills and Modalities:

\textbf{General Visual Perception}
\begin{itemize}
  \item A1. Detailed Image Description
  \item A2. Medium Image Description
  \item A3. Short Image Description
\end{itemize}

\textbf{Audio Perception}
\begin{itemize}
  \item B1. Video Subtitle Extraction
  \item B2. Audio Description
  \item B3. Music Description
\end{itemize}

\textbf{3D Visual Understanding}
\begin{itemize}
  \item C1. 3D Scene Description
  \item C2. 3D Situated Context Description
\end{itemize}

\textbf{Medical Visual Understanding}
\begin{itemize}
  \item D1. CT Scan Interpretation
  \item D2. Medical Image Description
\end{itemize}

\textbf{OCR/Text Extraction}
\begin{itemize}
  \item E1. General OCR
  \item E2. Poster/Slides Caption
  \item E3. PDF Text Extraction
\end{itemize}

\textbf{Structured Visual Data Interpretation }
\begin{itemize}
  \item F1. Chart/Plot Description
  \item F2. Table Description
\end{itemize}

\textbf{Mathematics and Geometry Extraction}
\begin{itemize}
  \item G1. Equation (LaTeX format)
  \item G2. Mathematics \& Geometry (LaTeX format)
\end{itemize}

\textbf{Instructions:}
\begin{enumerate}
  \item Only select skill/modality IDs necessary to answer the provided question.
  \item Respond strictly with the selected skill IDs, separated by commas.
\end{enumerate}

\textbf{Selected IDs:}  

\end{tcolorbox}
\caption{Prompts for expert selection.}
\label{fig:expert-prompt}
\end{figure*}

\end{document}